\documentclass[letterpaper]{article}
\usepackage[compat2,dvips,nohead]{geometry}
\usepackage{graphicx}
\usepackage{psfrag}
\usepackage{amsmath}
\usepackage{amsthm}
\usepackage{amssymb}
\usepackage{natbib}
\usepackage{url}
\usepackage{mathrsfs}
\usepackage{fltpage}            

\newcommand{\I}{\ensuremath{\mathbf{I}}}

\newcommand{\M}{\ensuremath{\mathbf{M}}}

\newcommand{\Q}{\ensuremath{\mathbf{Q}}}

\newcommand{\T}{\ensuremath{\mathbf{T}}}

\renewcommand{\aa}{\ensuremath{\mathbf{a}}}

\newcommand{\f}{\ensuremath{\mathbf{f}}}
\newcommand{\g}{\ensuremath{\mathbf{g}}}
\newcommand{\h}{\ensuremath{\mathbf{h}}}

\renewcommand{\t}{\ensuremath{\mathbf{t}}}

\newcommand{\x}{\ensuremath{\mathbf{x}}}
\newcommand{\y}{\ensuremath{\mathbf{y}}}
\newcommand{\z}{\ensuremath{\mathbf{z}}}
\newcommand{\0}{\ensuremath{\mathbf{0}}}

\newcommand{\btheta}{\ensuremath{\boldsymbol{\theta}}}

\newcommand{\bPsi}{\ensuremath{\boldsymbol{\Psi}}}
\newcommand{\bSigma}{\ensuremath{\boldsymbol{\Sigma}}}
\newcommand{\bTheta}{\ensuremath{\boldsymbol{\Theta}}}

\newcommand{\bbR}{\ensuremath{\mathbb{R}}}

\newcommand{\calA}{\ensuremath{\mathcal{A}}}
\newcommand{\calB}{\ensuremath{\mathcal{B}}}
\newcommand{\calC}{\ensuremath{\mathcal{C}}}

\newcommand{\calM}{\ensuremath{\mathcal{M}}}
\newcommand{\calN}{\ensuremath{\mathcal{N}}}
\newcommand{\calO}{\ensuremath{\mathcal{O}}}
\newcommand{\calP}{\ensuremath{\mathcal{P}}}

\newcommand{\calS}{\ensuremath{\mathcal{S}}}
\newcommand{\calT}{\ensuremath{\mathcal{T}}}

\newcommand{\calX}{\ensuremath{\mathcal{X}}}
\newcommand{\calY}{\ensuremath{\mathcal{Y}}}

\newcommand{\bydef}{\stackrel{\mathrm{\scriptscriptstyle def}}{=}}

\newcommand{\norm}[1]{\left\lVert#1\right\rVert}

\newcommand{\caja}[4][1]{{%
    \renewcommand{\arraystretch}{#1}%
    \begin{tabular}[#2]{@{}#3@{}}%
      #4%
    \end{tabular}%
    }}

{%
\begin{list}{#1}{
\vspace{-\topsep}
\vspace{-\partopsep}
\setlength{\itemindent}{0cm}
\setlength{\rightmargin}{0cm}
\setlength{\listparindent}{0cm}
\settowidth{\labelwidth}{#1}
\setlength{\leftmargin}{\labelwidth}
\addtolength{\leftmargin}{\labelsep}
\setlength{\itemsep}{0cm}
}%
}%
{%
\end{list}
\vspace{-\topsep}
\vspace{-\partopsep}
}

{\begin{enumerate}%
}%
{\end{enumerate}}

\DeclareGraphicsExtensions{.eps}
\graphicspath{{grf/}}

\bibpunct[, ]{(}{)}{;}{a}{,}{,} 

\title{Reconstruction of sequential data with density models\thanks{This paper was written on January 27th, 2004 and has not been updated since then.}} 
\author{Miguel {\'A}.\ Carreira-Perpi{\~n}{\'a}n \\
  Dept.\ of Computer Science, University of Toronto \\
  6 King's College Road. Toronto, ON M5S 3H5, Canada \\
  Email: \texttt{miguel@cs.toronto.edu}}
\date{}

\begin{document}

\maketitle

\begin{abstract}
  We introduce the problem of reconstructing a sequence of multidimensional real vectors where some of the data are missing. This problem contains regression and mapping inversion as particular cases where the pattern of missing data is independent of the sequence index. The problem is hard because it involves possibly multivalued mappings at each vector in the sequence, where the missing variables can take more than one value given the present variables; and the set of missing variables can vary from one vector to the next. To solve this problem, we propose an algorithm based on two redundancy assumptions: vector redundancy (the data live in a low-dimensional manifold), so that the present variables constrain the missing ones; and sequence redundancy (e.g.\ continuity), so that consecutive vectors constrain each other. We capture the low-dimensional nature of the data in a probabilistic way with a joint density model, here the generative topographic mapping, which results in a Gaussian mixture. Candidate reconstructions at each vector are obtained as all the modes of the conditional distribution of missing variables given present variables. The reconstructed sequence is obtained by minimising a global constraint, here the sequence length, by dynamic programming. We present experimental results for a toy problem and for inverse kinematics of a robot arm.

  \vspace*{1ex}\noindent\textbf{Keywords:} missing data reconstruction, multivalued (one-to-many) mappings, mapping inversion, Gaussian mixture modes, constraint minimisation, dynamic programming.
\end{abstract}

\section{Introduction}
\label{s:intro}

We consider the problem of reconstructing a sequence of multidimensional real vectors where some components of some vectors are missing. As an example, consider a speech utterance that is corrupted by another sound signal: at a given instant some speech spectral bands are corrupted by the other signal and can be considered missing. A given spectral band may be missing at some instants and present at other instants. The reconstruction problem here is to obtain the value of the missing spectral bands given the present ones, for the whole utterance. In the particular case when the components or variables that are missing are the same for all vectors of the sequence, the reconstruction problem becomes a regression or mapping approximation (of the missing variables given the present ones, or the Ys given the Xs).

Regression methods (e.g.\ a neural net) usually estimate univalued mappings (one-to-one or many-to-one), where a given value of the present variables results in a unique value for the missing variables. This works well when there is an underlying function (in its mathematical sense) that uniquely determines the missing variables given the present ones; but this is not the case generally, as for inverse mappings (resulting from inverting a forward, univalued map). For example, the position of the end-effector of a robot arm is uniquely determined by the angles at its joints, but not vice versa (inverse kinematics). When the missing data pattern varies along the sequence, univalued mappings will occur intertwined with multivalued mappings (one-to-many). We depart from the traditional point of view and define multivalued mappings as the basis of our reconstruction method: in one particular vector of the sequence, there may be more than one value for its missing components that is compatible with the value of its present components. A flexible way of generating such multivalued mappings (of an arbitrary subset of variables onto another subset) is via conditional distributions of a joint density model. However, some of these values will not be compatible with the rest of the sequence, considered globally: for example, they may result in physically impossible discontinuities of the inverse kinematics of the robot arm. To break the ambiguity of which reconstructed value to choose at each vector, we assume some prior information is available that constrains the sequence, in particular its continuity: one must choose the reconstructed values such that the resulting sequence is as continuous as possible. Thus, we assume that the data have two kinds of redundancy: (1) the vectors lie in a low-dimensional manifold and so knowledge of some vector components constrains the remaining ones (pointwise redundancy); and (2) consecutive vectors lie near each other (across-the-sequence redundancy).

The rest of the paper is organised as follows. Section~\ref{s:def} defines the data reconstruction problem. Section~\ref{s:deriv-funct-rel} explains how we derive multivalued functional relationships from conditional distributions. Section~\ref{s:constr} explains how to use prior information to constrain the multivalued mappings thus derived, and how to find the optimal reconstruction. Section~\ref{s:method} summarises the reconstruction method. Sections~\ref{s:exp-toy}--\ref{s:exp-robotarm} give experimental results. The remaining sections discuss the method, review related work and conclude the paper.

A preliminary version of this work appeared as a conference publication \citep{Carreir00a}.

\section{Definition of the problem of data reconstruction}
\label{s:def}

Generally, we define the problem of data reconstruction as follows: \emph{given a data set $\{\t_n\}^N_{n=1} \subset \bbR^D$ where part of the data are missing, reconstruct the whole data set to minimise an error criterion}. Let us examine in detail the elements of this definition.

\paragraph{The data set}

The \emph{data set} must have some structure in order that reconstruction be possible, i.e., there must be some dependencies between the different vectors in the set that give rise to redundancy. A typical example is sequential data in which consecutive vectors are close to each other (of course, not all sequences satisfy this). Such data can be considered as the result of sampling in time a vector that is a continuous function of the time. We can generalise this notion as follows. Assume $\{\t_n\}^N_{n=1}$ are samples from a continuous function $\mathfrak{F}$ of an independent variable \z\ at points $\{\z_n\}^N_{n=1}$. We call \z\ the \emph{experimental conditions}; \z\ can be the time (when the sample was taken), the position in space (where it was taken), etc. Thus, $\t = \mathfrak{F}(\z)$ is the sample point obtained at condition \z. We call $\t = (t_1,\dots,t_D)^T$ the \emph{observed variables}. We observe \t\ but not necessarily \z, and $\mathfrak{F}$ is unknown. Table~\ref{t:def:ex} gives some examples. $\mathfrak{F}$ gives to the collection $\{\t_n\}^N_{n=1}$ the structure or redundancy that allows the reconstruction of missing data.

\begin{table}
  \begin{center}
    \begin{tabular}{|l|l|l|}
\hline
\multicolumn{1}{|c}{Examples of problem} & \multicolumn{1}{|c}{Experimental conditions \z} & \multicolumn{1}{|c|}{Observed variables \t} \\
\hline
Trajectory of a mobile point & time, 1D & \caja[0.9]{t}{l}{spatial coordinates, 3D} \\
Spoken utterance & time, 1D & speech feature vector, $\approx$13D \\
Wind field on the ocean surface & spatial coordinates, 2D & wind velocity vector, 2D \\
Colour image & spatial coordinates, 2D & RGB level, 3D \\
\hline
    \end{tabular}
    \caption{Experimental conditions \z\ and observed variables \t\ for several examples of problems.}
    \label{t:def:ex}
  \end{center}
\end{table}

\begin{figure}
  \begin{center}
    \psfrag{T}[][]{$\calT$ (dimension $D = 3$)}
    \psfrag{M}[][]{$\calM$ (dimension $L = 2$)}
    \psfrag{F}[][]{$\mathfrak{F}(\z)$ (dimension $C = 1$)}
    \psfrag{t}[][]{$\t_n$}
    \includegraphics[width=\textwidth]{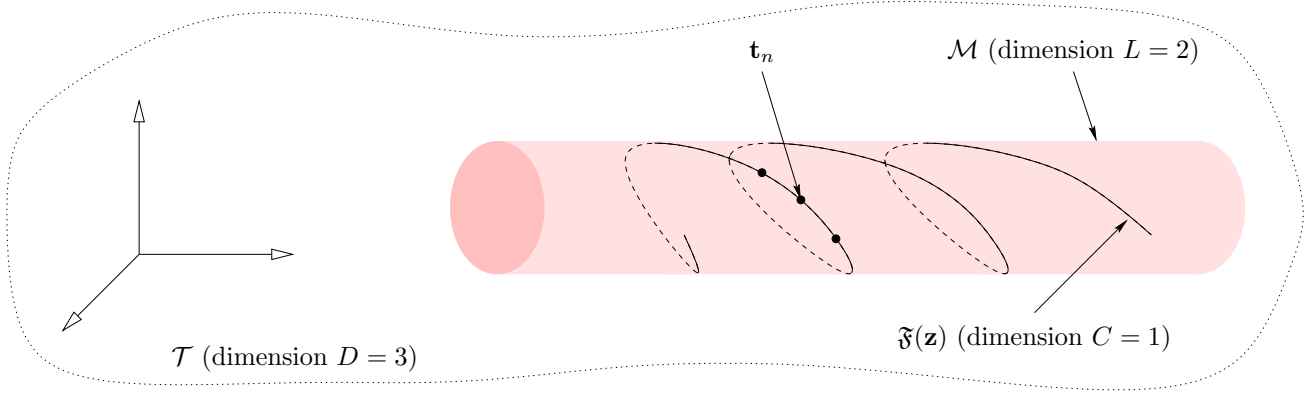}
    \caption{Dimensionalities involved in data reconstruction. The data, measured in a $D$-dimensional space \calT, live in an $L$-dimensional manifold \calM. A particular data set $\{\mathfrak{F}(\z)\}$ of \calM\ has a dimensionality $C$ equal to that of the experimental conditions \z. The dimensionalities verify $D \ge L \ge C$.}
    \label{f:def:dim}
  \end{center}
\end{figure}

We now have three dimensionalities: the dimensionality of the space \calT\ of the \t\ vectors that we observe, $D$; the intrinsic dimensionality of the manifold \calM\ of \calT\ where the \t\ vectors are constrained to live, $L$; and the dimensionality of the variable \z, $C$, corresponding to the batch of data $\{\t_n\}^N_{n=1}$ that we want to reconstruct. These dimensionalities verify $D \ge L \ge C$. For example (fig.~\ref{f:def:dim}), imagine an ant that walks on the trunk of a tree ($L=2$) and take \calT\ as the Euclidean space ($D=3$) and \z\ as the time ($C=1$); a given trajectory of the ant will be 1D, but the region that the ant is allowed to be on (the trunk) is 2D and in principle we may find it anywhere in that 2D region (either by taking a very long trajectory or by taking many different trajectories).

In the rest of this paper we assume $D > 1$. The case $D = 1$ does not allow to look at the relationship between variables, since there is only one ($\t = t_1$), and therefore we cannot make use of the redundancy derived from a low-dimensional representation. We will also assume sequential data unless otherwise noted, i.e., \z\ is the time (or some other 1D variable), although the treatment can be generalised to any dimensionality $C$. We will write $\{\t^{(n)}\}^N_{n=1}$ to denote a sequential data set, where $n$ indicates a temporal order in the data, reserving the notation $\{\t_n\}^N_{n=1}$ for data sets which need not have any sequential order (in which case $n$ is just a label).

\paragraph{The pattern of missing data}
\label{s:def:pat-md}

That \emph{part of the data are missing} means that some of the $ND$ variables $\{t_{nd}\}^{N,D}_{n,d=1}$ have missing values. We say that the value of a given scalar variable $t_{nd}$ is \emph{present} if such value was measured; otherwise, we say it is \emph{missing}. Abusing the language, we will also speak of present (missing) variables to mean variables whose values are present (missing). The reasons for a value to be missing are multiple: the value may not have been measured; the value was measured but may have got lost, erased or corrupted; and so on, depending on the particular problem.

When the values are corrupted in various amounts rather than just being either missing or present, one could consider further categories of uncertainty in a value. This can be a beneficial strategy for, e.g., recognition of occluded speech \citep{Cooke_01a}. However, for the purposes of data reconstruction, it is not clear what one should do with a value that is neither present (which must not be modified) nor missing (which must be filled in). Therefore we will stick to the present/missing dichotomy. We will also assume that each value has been classified as either present or missing, even though in some applications this task may not be trivial.

\begin{figure}
  \begin{center}
    \psfrag{t1}[]{\small $t_1$}
    \psfrag{t2}[]{\small $t_2$}
    \psfrag{t3}[]{\small $t_3$}
    \psfrag{t4}[]{\small $t_4$}
    \psfrag{t5}[]{\small $\cdots$}
    \psfrag{tD}[]{\small $t_D$}
    \psfrag{tt1}[]{\small $\t^{(1)}$}
    \psfrag{tt2}[]{\small $\t^{(2)}$}
    \psfrag{tt3}[]{\small $\t^{(3)}$}
    \psfrag{tt4}[]{\small $\t^{(4)}$}
    \psfrag{tt5}[]{\small $\t^{(5)}$}
    \psfrag{tt6}[]{\small $\t^{(6)}$}
    \psfrag{tt7}[]{\small $\cdots$}
    \psfrag{ttN}[]{\small $\t^{(N)}$}
    \psfrag{original}[][]{Original sequence}
    \psfrag{mask}[][]{Missing data pattern (mask)}
    \psfrag{observed}[][]{Observed sequence}
    \psfrag{variables}[b][]{\scriptsize Observed variables}
    \psfrag{vectors}[b][]{\scriptsize Vectors}
    \psfrag{present}[b][t]{\scriptsize Present value}
    \psfrag{missing}{\scriptsize Missing value}
    \includegraphics[width=\textwidth]{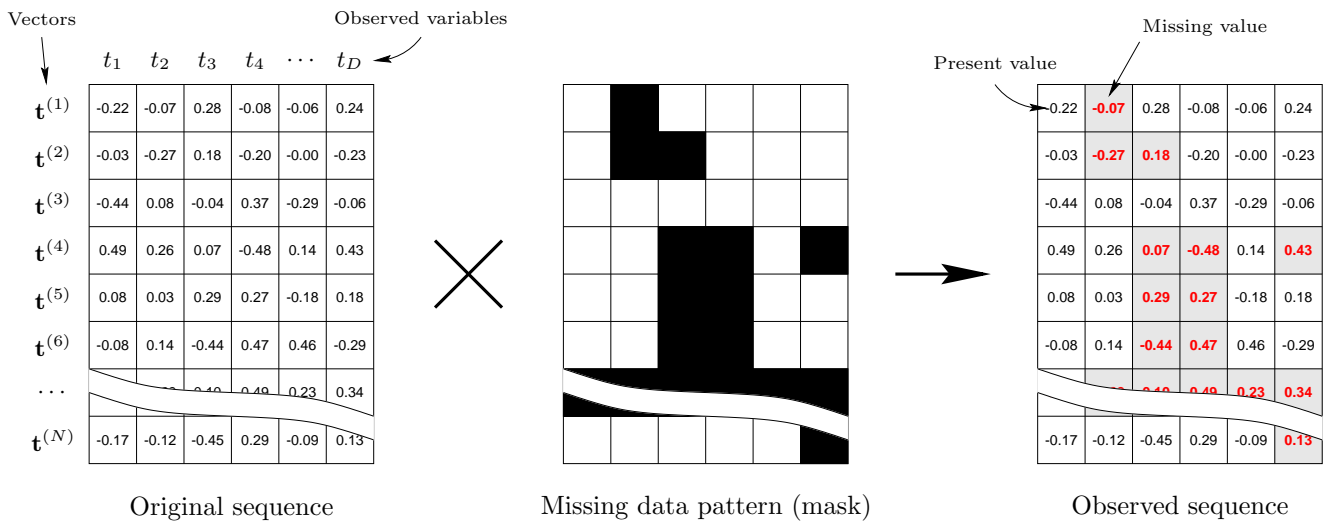}
    \caption{Schematic representation of the missing data. The black and white cells in the missing data pattern indicate missing and present values, respectively.}
    \label{f:def:mask}
  \end{center}
\end{figure}

We can represent the pattern of missing data associated with the data set $\{\t_n\}^N_{n=1}$ with a matrix (or \emph{mask}) $\M = (m_{nd})$ of $N \times D$ such that $m_{nd} = 1$ if the value of $t_{nd}$ is present and $m_{nd} = 0$ otherwise. The matrix \M\ acts then as a multiplicative mask on the complete data set, i.e., the data set where all values are present, as represented in figure~\ref{f:def:mask}. We obtain the problem of regression (or mapping approximation) in the particular case where $m_{nd}$ is independent of $n$ (in which case the mask of fig.~\ref{f:def:mask} has a columnar structure). We will use the term \emph{regression-type missing data pattern} if the pattern of missing data is constant over the sequence and \emph{general missing data pattern} if it varies. We will use the term \emph{complete data set} to mean the data set as if no values were missing and \emph{reconstructed data set} to mean the data set where the missing values have been filled in by a reconstruction algorithm.

The algorithm described in this paper is applicable to any pattern of missing data, irrespectively of why or how that pattern came about. However, if the missing data are not missing completely at random (i.e., the pattern of missing data depends on the actual data; \citealp{LittleRubin87}), then any information about the mechanism of missing data should be taken into account if possible, since it may further constrain the values that the missing variables may take.

\paragraph{Reconstruction of the whole data set}

\emph{To reconstruct the whole data set} means to find a single estimate of each missing value. Generally, we define four types of reconstruction according to the combinations of the following characteristics:
\begin{itemize}
\item The number of candidate reconstructions of a given entity that are provided: \emph{single} or \emph{multiple} reconstruction.
\item The entity that is being reconstructed: \emph{pointwise} (or \emph{local)} for reconstruction of a vector $\t^{(n)}$ given information present in $\t^{(n)}$ only, and \emph{global} for reconstruction of the whole sequence or data set $\{\t^{(n)}\}^N_{n=1}$ given information present in it.
\end{itemize}
Using this terminology, we seek a single global reconstruction of the data set. A method that provides single pointwise reconstruction can only provide single global reconstruction; standard regression and mapping approximation are examples of such methods. But single global reconstruction can be attained by an appropriate combination of a collection of multiple pointwise reconstructions; the method proposed in this paper does this.

\paragraph{Error criterion}

In this paper we use the square difference between the true and the reconstructed vectors (averaged over the sequence) as a measure of the reconstruction error. Other criteria are also possible.

\paragraph{Notation}

We use the following notation to select components (variables) of a vector. If $\t = (t_1,\dots,t_D)^T \in \calT$ is a $D$-dimensional real vector and $\calP \subset \{1,\dots,D\}$ is a set of indices, then $\t_{\calP}$ represents the vector composed of those components of \t\ whose indices are in \calP. For example, if $\calP=\{1,7,3\}$ and $\calM=\{2,5\}$ then $p(\t_{\calM}|\t_{\calP})$ is $p(t_2,t_5|t_1,t_3,t_7)$. This notation is convenient to represent arbitrary patterns of missing data, where the present or missing variables at point $n$ are indicated by sets $\calP_n$ and $\calM_n$ satisfying $\calP_n \cap \calM_n = \varnothing$ and $\calP_n \cup \calM_n = \{1,\dots,D\}$. Abusing the notation, we may sometimes write $\t_{n,\calP}$ or $\t^{(n)}_{\calP}$ to mean $\t_{n,\calP_n}$ or $\t^{(n)}_{\calP^{(n)}}$, respectively. Often, we will also use \x\ and \y\ to refer to the present and missing variables, respectively.

\section{Deriving multivalued functional relationships from conditional distributions}
\label{s:deriv-funct-rel}

Any kind of reconstruction is based on a functional relationship of what is missing given what is present. This section discusses two central ideas. The first one is that one can define a functional relationship $\x \rightarrow \y$ (\y\ as a function of \x) from the conditional distribution of \y\ given \x\ by picking representative points of this distribution. The second one is that underlying a multimodal conditional distribution there (often) is a multivalued functional relationship and that it is wrong to summarise such a distribution with its expected value. Instead, we propose the use of all the modes of a conditional distribution to define a multivalued functional relationship and thus to define multiple pointwise reconstruction. We assume that the conditional distribution comes from a probability density function (pdf) $p(\t)$ for all the observed variables $\t = (t_1,\dots,t_D)^T$.

In general, we use the terms \emph{predictive distribution} for a distribution containing information about the missing variables (perhaps different from the conditional) and \emph{candidate (pointwise) reconstructions} for the values used to fill in the missing variables (perhaps different from the modes).

\subsection{Informative, or sparse, distributions}
\label{s:deriv-funct-rel:sparse}

\begin{figure}
  \begin{center}
    \psfrag{x}[]{$y$}
    \psfrag{y}[]{$x$}
    \psfrag{x0}[]{$y_0$}
    \psfrag{x1}[]{$y_1$}
    \psfrag{x2}[]{$y_2$}
    \psfrag{y0}[]{$x_0$}
    \psfrag{px_y}[][Bl][1][90]{$p(y|x=x_0)$}
    \psfrag{pxy}[b]{$p(x,y)$}
    \includegraphics[width=.8\textwidth]{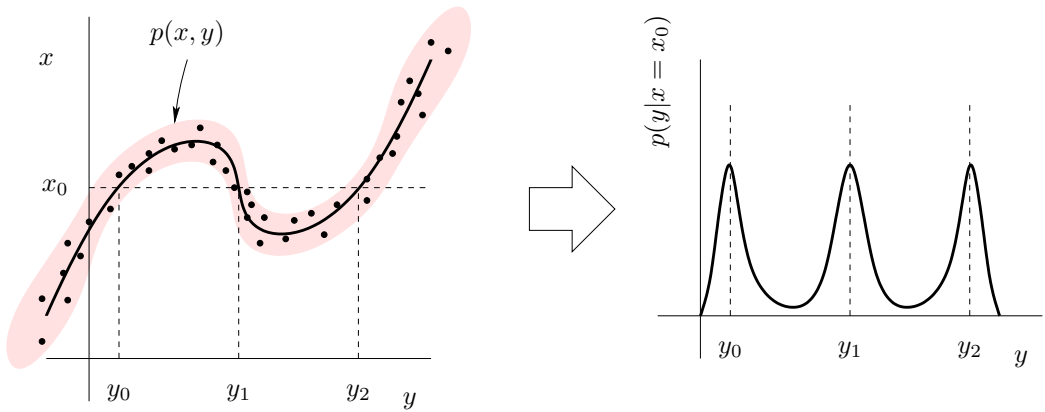}
    \caption{A conditional distribution can imply a multivalued functional relationship. \emph{Left}: joint probability density $p(x,y)$, represented by the shaded area; the black dots show a sample typical from that density and the thick line indicates a low-dimensional manifold. \emph{Right}: multivalued mapping $x \rightarrow y$ from the multimodal conditional distribution $p(y|x)$. For $x = x_0$ we obtain $y_0$, $y_1$ and $y_2$ as possible values for $y$.}
    \label{f:deriv-funct-rel:sparse}
  \end{center}
\end{figure}

A conditional distribution $p(y|x)$ which consists of several sharp peaks conveys information about a functional relationship in that the probability mass is concentrated around a few points. We can construct a multivalued mapping $x \rightarrow y$ by picking several particularly informative points of the domain of $p(y|x)$ (see fig.~\ref{f:deriv-funct-rel:sparse}). In general, we say that a $D$-dimensional pdf $p(\t)$ (possibly the distribution of \t\ conditioned on the values of some other variables) is \emph{informative} or \emph{sparse} if the probability mass is concentrated around a low-dimensional manifold. Conversely, if the probability mass is spread over a $D$-dimensional region then we say that the pdf is \emph{uninformative}. We thus state the principle that \emph{highly informative distributions can be assimilated to (possibly multivalued) functional relationships}.

As we define it, the concept of distribution sparseness is a probabilistic extension of the concept of low-dimensional manifold. Thus, a $D$-dimensional pdf defined around a point/curve/surface is sparse for $D \ge 1/2/3$, respectively, and so on. Our definition of sparseness is vague, since the term ``concentrated around'' is relative---just how much probability mass and how near the low-dimensional manifold? For the purpose of deriving functional relationships from conditional distributions this definition suffices. \citet{Carreir01a} discusses the issue of measuring sparseness quantitatively.

A conditional distribution $p(y|x)$ may be informative for some values of $x$ and uninformative for other values of $x$. If we require that $p(y|x)$ be informative for any value of $x$, then the joint density $p(x,y)$ must be itself an informative distribution, since $p(y|x) \propto p(x,y)$.

\subsection{Unimodal distributions: $L_2$-optimality of the mean}
\label{s:deriv-funct-rel:mean}

For unimodal distributions, it makes sense to summarise the distribution with a single point. Which point to choose depends on what error criterion we want to minimise. If we pick the point that minimises the average distance (with respect to the distribution of \t) to every other point in the domain of \t, $\hat{\t} = \arg\min_{\hat{\t} \in \calT}{\int_{\calT}{p(\t) d(\t,\hat{\t}) \, d\t}}$, then the optimal point is the mean of $p(\t)$ if $d$ is given by the $L_2$-norm \citep[pp. 201--205]{Bishop95a} and the median if $d$ is given by the $L_1$-norm \citep[p.~209--210]{Degroot86a}. For symmetric distributions the mean, median and mode coincide, but for skewed distributions they differ. However, the median does not have a natural generalisation to more than 1D. A number of approaches exist that derive univalued mappings from the conditional distribution, usually via the mean (see section~\ref{s:rel-work:multiv-map}).

\subsection{Multimodal distributions: the mean considered harmful}
\label{s:deriv-funct-rel:mean-harm}

The mean of a multimodal distribution can lie on an area of low probability, thus being a highly unlikely representative of the distribution. Worse still, it may lie outside of the support of the random variable when such support is not a convex set, since the mean is a convex sum itself (and this can happen even if the distribution is unimodal); this fact has been pointed out in the context of inverse problems \citep{Jordan90,JordanRumelh92}. In spite of this, the mean remains the most common choice of representative point of a multimodal distribution, possibly due to being optimal with respect to the $L_2$-norm (as long as one is constrained to choose a single point) and to its ease of calculation.

A better choice are the modes. Unlike the mean, any mode by definition must lie inside the support of the random variable and have a locally high probability density. In general, calculating the modes of a multivariate distribution is difficult because one does not know how many modes there are and because computing each one of them cannot be done analytically. However, for Gaussian mixtures, \citet{Carreir00b} gives algorithms for finding all the modes%
\footnote{The algorithms work by starting a hill-climbing search from every centroid. A Gaussian mixture in $2$ or more dimensions can have more modes than components even if all the components have the same, isotropic covariance \citep{CarreirWilliam03b,CarreirWilliam03c}, so the algorithms can miss some modes. However, this situation is very infrequent.}.
The algorithms can also compute error bars for each mode, thus estimating locally the error, though this will not be used here.

But then how does one deal with a multivalued choice? In the absence of additional information, all we can do is to keep all the modes, since any of them is a likely representative of the distribution. This implies defining a multivalued functional relationship. Thus, we propose to select \emph{all the modes of the conditional distribution} as representative points of it. 

\subsection{Sampling the predictive distribution}
\label{s:deriv-funct-rel:sampling}

Another method to pick representative points of a distribution is to sample from it. This makes sense when there are few present variables, so that the missing variables are so underdetermined that they can take values on a continuous manifold: in this case, the modes of a conditional distribution are effectively a particular sample from that manifold. Computationally, sampling can also be more attractive than finding all the modes.

However, when the missing variables are constrained to take a finite number of values, this does not make much sense. Unless the conditional distribution is very sharply peaked around its modes, sampling will return values that by definition are corrupted by noise: sometimes they may fall in areas far from the main probability mass body or ignore low-probability bumps (which may represent infrequent but legitimate reconstructions). A further, serious problem is how to set the number of samples to obtain, which will certainly locally underestimate or overestimate the true number of values that the missing variables can take. Missing a correct pointwise reconstruction or generating a wrong one may affect the global reconstruction, not just the local one, via the continuity constraint (see section~\ref{s:constr}). Our experiments show that the sampling strategy performs consistently worse than both the mean- and mode-based approaches.

\subsection{Joint density model of the observed variables}
\label{s:deriv-funct-rel:density}

For a generic missing data pattern we need to be able to obtain the conditional distribution for any combination of missing and present variables, which requires estimating a joint density model of all the observed variables. The joint density embodies the relation of any subset of variables with any other subset of variables; all we need is to compute the appropriate conditional distribution, which in turn requires a marginalisation: $p(\t_{\calM} | \t_{\calP}) = p(\t_{\calM}, \t_{\calP}) / p(\t_{\calP}) = p(\t) / p(\t_{\calP})$. Therefore, we are free to choose the method by which we estimate the joint density as long as the estimator allows easy marginalisation. The density model should be estimated offline using a training set different from the one that is to be reconstructed. Typically, the training set will have no missing data, although even if it does, it is possible to train the model using an EM algorithm (e.g.\ \citealp{GhahramJordan94b,MclachKrishn97a}).

A suitable class of density models are continuous latent variable models, since the pointwise redundancy implies an intrinsic low dimensionality of the observed variables \citep{Carreir01a}. The distribution of the observed variables is sparse in the sense of section~\ref{s:deriv-funct-rel:sparse}. The density model must be able to represent multimodal densities. This discards linear-normal models (factor analysis and principal component analysis). It should also allow an easy computation of conditional distributions. Useful models include the generative topographic mapping (GTM; \citealp*{Bishop_98a}), mixtures of factor analysers \citep{GhahramHinton96a} and mixtures of principal component analysers \citep{TippinBishop99a}. For all these models the density of the observed variables has the form of a (constrained) Gaussian mixture and the number of tunable parameters can be kept reasonably low while keeping the ability to represent a broad class of multimodal densities. We can also directly use Gaussian mixtures or (nonparametric) kernel density estimation, both of which have the property of universal density approximation \citep{Titter_85a,Scott92}. In this paper we use the GTM; we refer the reader to the original papers \citep{Bishop_98a,Bishop_98b} for details on this model.

Hereafter we will assume that the joint density has the form of a Gaussian mixture, whose parameters were estimated from a data sample in an unsupervised way. Computing conditional distributions is then straightforward (for a diagonal Gaussian mixture, this requires little more than crossing out rows and columns). Finding all the modes can be done with the algorithms of \citet{Carreir00b}. In the particular case where the pattern of missing data is constant, one can just model the appropriate conditional distribution rather than the joint density, of course (section~\ref{s:rel-work:cond}).

\section{Use of prior information to constrain multivalued mappings}
\label{s:constr}

So far we have exploited the redundancy between component variables of a given data point (via the conditional distribution) to constrain the possible values of the missing variables, but this can still result in multivalued mappings, as we have seen. In the absence of any additional information, the answer to the reconstruction problem would be those multivalued mappings. We now turn to the case of using extra information about the problem to constrain the possible values so that we obtain a unique global reconstruction of a data set.

\subsection{Continuity constraints}
\label{s:constr:cont}

For many practical cases, additional information comes from the redundancy between data points and depends on the experimental conditions. The most usual such constraint is given by \emph{continuity} of the variables as a function of the experimental conditions, $\t = \mathfrak{F}(\z)$: nearby values of \z\ produce nearby values of \t, or ``$d(\t_n,\t_{n+1})$ is small'' according to a distance dependent on the problem. In this case, typically \z\ is a physical variable such as the time (or space) and then \t\ is a measured vector that depends continuously on it, resulting in a continuous trajectory (or field). In general, we can define constraints on the trajectory by bounding via a norm the derivatives of the function $\mathfrak{F}$ (if they exist), numerically approximating each derivative by a finite-difference scheme in terms of the available measures at $z_1, z_2,\dots, z_N$ and applying suitable boundary conditions at the trajectory ends (e.g.\ to consider open or closed trajectories). \emph{Continuity} (which penalises sharp jumps) results from the first derivative, while \emph{smoothness} (which penalises sharp turns) results from the second. This results in the norm of a linear combination of points near $\t_n$. Actual examples are given in section~\ref{s:constr:min}.

The norm depends on the problem, but typically will be the Euclidean or squared Euclidean norm. If different variables have different units or scales it may be convenient to use a weighted norm. The squared Euclidean norm has the computational advantage that it separates additively along each variable and so for a constant missing data pattern ($\calM_n = \calM \ \forall n$) we need only consider the missing variables (since the present ones contribute a constant additive term). It also results in constraints that are quadratic forms on the variables $\{t_{nd}\}^{N,D}_{n,d=1}$, though this makes no difference in the constraint minimisation.

Another type of constraint that has often been found useful in inverse problems is a \emph{quadratic} constraint $(\t_{n}-\t_0)^T \Q (\t_{n}-\t_0)$, which can be interpreted physically as an energy in mechanical systems. In particular, $\norm{\t_{n}-\t_0}^2$ corresponds to the potential energy of a harmonic oscillator with resting position at $\t = \t_0$ and restoring constant $k = 2$.

\subsection{Constraint by forward mapping}
\label{s:constr:fwd-map}

In the particular case where the reconstruction problem is a mapping inversion problem, we can use the known forward (direct) mapping as a further constraint. The forward mapping \g\ maps the missing variables onto the present ones: $\t_{\calP} = \g(\t_{\calM})$, where \calP\ and \calM\ are independent of $n$ (i.e., the same for all data points). Thus, given the values of $\t_{\calP}$ and given a candidate reconstruction $\hat{\t}_{\calM}$ (for example $\hat{\t}_{\calM}$ could be one of the modes of $p(\t_{\calM}|\t_{\calP})$), then the distance between $(\t_{\calP} \ \hat{\t}_{\calM})$ and $(\g(\hat{\t}_{\calM}) \ \hat{\t}_{\calM})$ should be as small as possible.

In the ideal case where the procedure that provides candidate reconstructions (in this paper, the modes of the conditional distribution) was perfect, this constraint would have no effect, since every $\hat{\t}_{\calM}$ would exactly map onto $\t_{\calP}$. In reality, correct reconstructions will give small, but nonzero, differences between $\t_{\calP}$ and $\g(\hat{\t}_{\calM})$, while incorrect reconstructions (such as spurious modes) will generally give a much larger difference. Thus, the constraint by forward mapping can help to discard such incorrect reconstructions.

\subsection{Minimisation of a global constraint}
\label{s:constr:min}

The constraints introduced above are by definition local and generally take the form of the (squared) norm of a linear combination of neighbouring points; e.g.\ $\norm{\t_{n+1} - \t_n}$. Now we derive from such local constraints a \emph{global constraint} that involves the whole data set. This way we define an objective function depending on all missing variables and then find the combination of candidate pointwise reconstructions that minimises it. We will then have a single global reconstruction that should be a good approximation to the complete data set. The role of the global constraint in breaking the nonuniqueness of the reconstruction is analogous to that of regularising operators for ill-posed problems \citep{TikhonArsenin77a}.

We can define a global constraint by adding the local constraints for each point in the sequence, thus obtaining:
\begin{align}
  \label{e:constr:min:cont}
  \text{Continuity, } \mathscr{C} &\bydef \sum^{N-1}_{n=1}{\norm{\t^{(n)} - \t^{(n+1)}}} \\
  \label{e:constr:min:smooth}
  \text{Smoothness, } \mathscr{S} &\bydef \sum^{N-1}_{n=2}{\norm{\t^{(n+1)} - 2\t^{(n)} + \t^{(n-1)}}} \\
  \label{e:constr:min:quad}
  \text{Quadratic, } \mathscr{Q} &\bydef \sum^N_{n=1}{(\t^{(n)}-\t_0)^T \Q (\t^{(n)}-\t_0)} \\
  \label{e:constr:min:fwd}
  \text{Forward-mapping, } \mathscr{F} &\bydef \sum^N_{n=1}{\norm{\left(\begin{smallmatrix} \t^{(n)}_{\calP} \\ \t^{(n)}_{\calM} \end{smallmatrix}\right) - \left(\begin{smallmatrix} \g(\t^{(n)}_{\calM}) \\ \t^{(n)}_{\calM} \end{smallmatrix}\right)}}.
\end{align}
In eqs.~\eqref{e:constr:min:cont}--\eqref{e:constr:min:smooth} we have used first- and second-order forward differences in $\mathscr{C}$ and $\mathscr{S}$, respectively, assuming that the experimental condition variable $z$ is sampled regularly; if this is not the case, then one should weight term $n$ by $1/(z_{n+1} - z_n)$. Note that $\mathscr{C}$ is the length of the polygonal trajectory passing through $\t^{(1)},\dots,\t^{(N)}$. We can define a global constraint as a linear combination of constraints such as those above, but in this paper we will concentrate in $\mathscr{C}$.

\begin{figure}
  \begin{center}
    \small
    \psfrag{B}[]{$B$}
    \psfrag{E}[]{$E$}
    \psfrag{n}[]{$n=$}
    \psfrag{n1}[]{$1$}
    \psfrag{n2}[]{$2$}
    \psfrag{n3}[]{$3$}
    \psfrag{n4}[]{$4$}
    \psfrag{n5}[]{$\cdots$}
    \psfrag{n6}[]{$N$}
    \psfrag{v}[]{$\nu_n =$}
    \psfrag{v1}[]{$5$}
    \psfrag{v2}[]{$2$}
    \psfrag{v3}[]{$4$}
    \psfrag{v4}[]{$1$}
    \psfrag{v5}[]{$\cdots$}
    \psfrag{v6}[]{$\nu_N$}
    \psfrag{possiblepaths}[r][Br]{\caja[1.5]{c}{c}{$\prod^N_{n=1}{\nu_n}$ \\ possible paths} $\left\{\rule{0pt}{1cm}\right.$}
    \psfrag{candidates}[l][Bl]{\caja{c}{c}{Candidates to \\ reconstruct $\t^{(n)}$}}
    \includegraphics[width=0.9\textwidth]{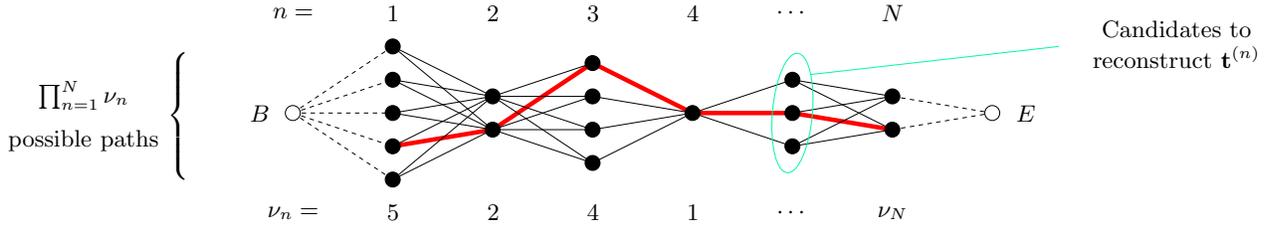}
    \caption{Constraint minimisation as a shortest path problem in a layered graph. There are $\nu_n$ nodes in layer $n$ in the graph, which are the candidate pointwise reconstructions of $\t^{(n)}_{\calM^{(n)}}$ (the modes of the conditional distribution $\t^{(n)}_{\calM^{(n)}}|\t^{(n)}_{\calP^{(n)}}$). There are $N$ layers and a single global reconstruction of the data set defines a left-right (or right-left) path passing through all layers exactly once (one such path is highlighted). By imagining fictitious end nodes $B$ and $E$ fully connected by zero-cost links to the end layers $1$ and $N$, respectively, we can reformulate the problem as that of finding the shortest path between $B$ and $E$. Each edge in the graph is undirected and has an associated cost given by the continuity constraint (the distance between the reconstructed points).}
    \label{f:constr:min}
  \end{center}
\end{figure}

The global constraint is a function of the missing variables $\{\t^{(n)}_{\calM^{(n)}}\}^N_{n=1}$. Thus we arrive at the following minimisation problem:
\begin{equation*}
  \text{Reconstruct the data set as } \arg\min_{\{\t^{(n)}_{\calM^{(n)}}\}^N_{n=1} \in \calS} {\mathscr{C}\left(\{\t^{(n)}_{\calM^{(n)}}\}^N_{n=1}\Big|\{\t^{(n)}_{\calP^{(n)}}\}^N_{n=1}\right)}
\end{equation*}
where the search space \calS\ is the Cartesian product of the $N$ sets of candidate reconstructions for each $\t^{(n)}_{\calM^{(n)}}$ (i.e., each set contains the modes of $p(\t^{(n)}_{\calM^{(n)}}|\t^{(n)}_{\calP^{(n)}})$). This is a combinatorial optimisation problem that can be expressed as finding the shortest path in a layered graph, as represented in figure~\ref{f:constr:min}. Calling $\nu_n \ge 1$ the number of candidate pointwise reconstructions at point $n$, the total number of paths is $\prod^N_{n=1}{\nu_n}$, which in an average case is of exponential order in $N$. Fortunately, there are efficient algorithms both for global (exact) and local (approximate) minimisation.

\subsection{Global minimisation: dynamic programming}
\label{s:constr:min-dp}

\begin{figure}
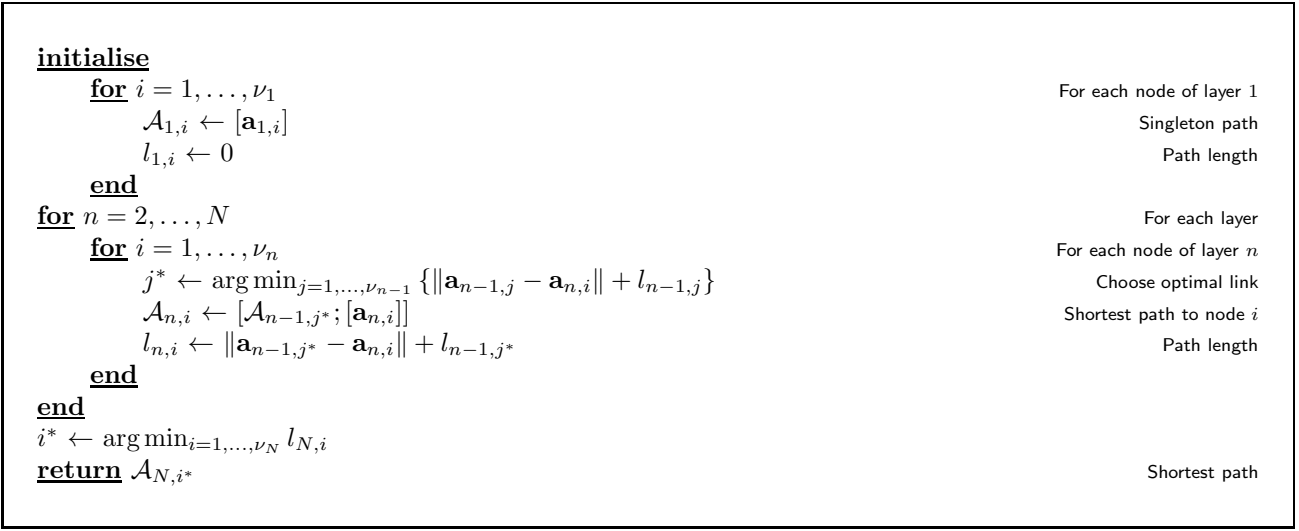

  \begin{center}
    \fbox{%
      \quad%
      \begin{minipage}[c]{.94\textwidth}
        \begin{tabbing}
          \\
          mm \= mm \= mm \= mm \= mm \= \kill
          \underline{\textbf{initialise}} \+ \\
          \underline{\textbf{for}} $i=1,\dots,\nu_1$ \` {\scriptsize \textsf{For each node of layer $1$}} \+ \\
          $\calA_{1,i}$ $\leftarrow$ $[\aa_{1,i}]$ \` {\scriptsize \textsf{Singleton path}} \\
          $l_{1,i}$ $\leftarrow$ $0$ \` {\scriptsize \textsf{Path length}} \- \\
          \underline{\textbf{end}} \- \\
          \underline{\textbf{for}} $n=2,\dots,N$ \` {\scriptsize \textsf{For each layer}} \+ \\
          \underline{\textbf{for}} $i=1,\dots,\nu_n$ \` {\scriptsize \textsf{For each node of layer $n$}} \+ \\
          $j^*$ $\leftarrow$ $\arg\min_{j=1,\dots,\nu_{n-1}}{\left\{\norm{\aa_{n-1,j} - \aa_{n,i}} + l_{n-1,j}\right\}}$ \` {\scriptsize \textsf{Choose optimal link}} \\
          $\calA_{n,i}$ $\leftarrow$ $[\calA_{n-1,j^*} ; [\aa_{n,i}]]$ \` {\scriptsize \textsf{Shortest path to node $i$}} \\
          $l_{n,i}$ $\leftarrow$ $\norm{\aa_{n-1,j^*} - \aa_{n,i}} + l_{n-1,j^*}$ \` {\scriptsize \textsf{Path length}} \- \\
          \underline{\textbf{end}} \- \\
          \underline{\textbf{end}} \\
          $i^*$ $\leftarrow$ $\arg\min_{i=1,\dots,\nu_N}{l_{N,i}}$ \\
          \underline{\textbf{return}} $\calA_{N,i^*}$ \` {\scriptsize \textsf{Shortest path}} \\
        \end{tabbing}
      \end{minipage}%
      \quad%
      }
    \caption{Dynamic programming algorithm for global constraint minimisation. Sequences of nodes are written in square brackets and $[\calA ; \calB]$ means the concatenation of sequences \calA\ and \calB.}
    \label{f:constr:min-dp}
  \end{center}
\end{figure}

The problem of finding the shortest path in a layered graph is a particular case of that of finding the shortest path in an acyclic graph and can be conveniently solved by dynamic programming \citep{Bellman57a,Bertsek87a}. We can apply dynamic programming because for this problem the following principle of optimality for a decision policy holds: \emph{regardless of the policy adopted at previous stages, the remaining decisions must constitute an optimal policy}, where here a \emph{stage} is a layer in the graph and a \emph{policy} is a sequence of decisions (i.e., a sequence of chosen nodes). This leads us to an algorithm where the decision of what link to choose is taken locally (at each layer $n$), but $\nu_n$ paths must be kept. Figure~\ref{f:constr:min-dp} gives the forward recursion version of the dynamic programming algorithm (starting from layer $1$) for a global continuity constraint $\mathscr{C}$. Due to the symmetry of the problem, a backward algorithm (starting from layer $N$) is equivalent. The algorithm requires the following definitions:
\begin{center}
  \begin{tabular}[c]{lp{0.8\textwidth}}
    $\{\aa_{n,i}\}^{\nu_n}_{i=1}$ & The set of candidate pointwise reconstructions for $\t^{(n)}$; thus $\aa_{n,i}$ is node $i$ of layer $n$ in the graph. \\
    $\calA_{n,i}$ & Minimal length path from layer $1$ to node $i$ of layer $n$, for $i=1,\dots,\nu_n$. Thus, $\calA_{n,i} = [\aa_{1,\bullet}; \aa_{2,\bullet}; \dots; \aa_{n,i}]$ where $\bullet$ indicates some node. \\
    $l_{n,i}$ & Total length of $\calA_{n,i}$, i.e., $l_{n,i} \bydef \sum^{n-1}_{m=1}{\norm{\calA_{n,i}(m) - \calA_{n,i}(m+1)}}$, where $\calA_{n,i}(m)$ is the $m$th element of the sequence $\calA_{n,i}$.
  \end{tabular}
\end{center}
We disregard the unlikely case of ties, where the $\arg\min_{j=1,\dots,\nu_{n-1}}$ operation may return several values of $j$.

The dynamic programming algorithm examines each link in the graph (i.e., each pair of nodes in adjacent layers) only once, thus achieving its mission very efficiently.

\subsection{Local minimisation: greedy algorithm}
\label{s:constr:min-greedy}

A more intuitive and slightly faster algorithm is obtained as a greedy version of dynamic programming: at layer $n$, it simply selects the minimal cost edge (i.e., the closest node). The starting point can be any node in any layer $n_0$, but to improve the chances of getting a good path, it is better to start in a layer having very few nodes (hopefully just one). The algorithm greedily proceeds from $n_0$ leftwards to $1$ and rightwards to $N$, since all edges are undirected. Unlike dynamic programming, this algorithm needs only keep a single path at any time rather than $\nu_n$ paths, but it does not necessarily find a path of globally minimal cost. Our experiments show that it usually leads to poor solutions, not just in terms of a high value of the constraint, but also as yielding a high reconstruction error of the dataset---which is our ultimate criterion. Such solutions are sensitive to the choice of starting layer $n_0$. Also, the greedy algorithm has a tendency to obtain reconstructed trajectories that retrace themselves at turning points and have abrupt jumps.

\section{Summary of the method}
\label{s:method}

We can now summarise our reconstruction method (see also fig.~\ref{f:concl}). The first step is done offline and involves estimating a Gaussian-mixture joint density model $p(\t)$ of the observed variables, using a complete dataset $\{\t_n\}^{N'}_{n=1}$ (we use GTM in this paper). At reconstruction time, we are given a sequence $\{\t^{(n)}\}^N_{n=1}$ with missing data. Then:
\begin{enumerate}
\item For each vector $\t^{(n)}$ in the sequence, $n = 1,\dots,N$:
  \begin{itemize}
  \item Compute the conditional distribution $p(\t^{(n)}_{\calM^{(n)}}|\t^{(n)}_{\calP^{(n)}})$ of the missing variables given the present ones. This is a Gaussian mixture too.
  \item Find all the modes of this conditional distribution. These are the candidate reconstructions for $\t^{(n)}$.
  \end{itemize}
\item Minimise the trajectory length $\mathscr{C}\left(\{\t^{(n)}\}^N_{n=1}\right)$ over the set of candidate reconstructions using dynamic programming to yield the reconstructed sequence.
\end{enumerate}

\section{Experiments: 2D toy example}
\label{s:exp-toy}

In this section we study the performance of the reconstruction method with a 2D toy problem with observed variables $\left(\begin{smallmatrix} t_1 \\ t_2 \end{smallmatrix}\right)$. The mapping $t_1 \rightarrow t_2$ is many-to-one and so easy to estimate by traditional methods, but the mapping $t_2 \rightarrow t_1$ is one-to-many, as is the mapping $\varnothing \rightarrow \left(\begin{smallmatrix} t_1 \\ t_2 \end{smallmatrix}\right)$.

We consider the forward, nonlinear mapping $g(x) \bydef x+3\sin{x}$ for $x\in[-2\pi,2\pi]$, which results in 1D data ($L=1$) observed in $D=2$ dimensions by taking $\t=\left(\begin{smallmatrix} t_1 \\ t_2 \end{smallmatrix}\right)$ with $t_1 = x$ and $t_2 = g(x)$; see fig.~\ref{f:exp-toy:results}A. The forward mapping $g$ is injective only in parts of the domain and so the inverse mapping $g^{-1}$ is sometimes multivalued. The task is to reconstruct a (possibly noisy) sampled trajectory of $N$ 2D points such as that of fig.~\ref{f:exp-toy:results}A with missing data. The reconstruction error is computed as the average squared error $\frac{1}{N}\sum^N_{n=1}{\norm{\t^{(n)}-\hat{\t}^{(n)}}^2}$, where $\{\t^{(n)}\}^N_{n=1}$ is the original, complete trajectory and $\{\hat{\t}^{(n)}\}^N_{n=1}$ the trajectory reconstructed by a particular method.

Five types of $N \times 2$ mask are considered:
\begin{itemize}
\item $\M_{\text{fwd}}$ where $t_2$ is always missing (regression $t_1 \rightarrow t_2$, $50$\% missing data);
\item $\M_{\text{inv}}$ where $t_1$ is always missing (regression $t_2 \rightarrow t_1$, $50$\% missing data);
\item $\M_{\text{$75$\%}}$, $\M_{\text{$50$\%}}$, $\M_{\text{$25$\%}}$ where any of $t_1$, $t_2$ are missing at random ($75$\%, $50$\%, $25$\% missing data, respectively).
\end{itemize}
$\M_{\text{fwd}}$ and $\M_{\text{inv}}$ are regression-type masks, while $\M_{\text{$75$\%}}$--$\M_{\text{$25$\%}}$ are general missing data patterns. By applying the mask to a complete trajectory, we obtain a trajectory with missing data (see fig.~\ref{f:def:mask}).

As training data for the joint density model $p(\t)$, we generated a shuffled (i.e., without sequential order) training set $\{\t_n\}^{N'}_{n=1}$ with $N'=1\,000$ points sampled from the curve with additive normal $(\0,\sigma^2\I)$ noise for $\sigma = 0.2$. We used it to train $3$ models%
\footnote{We gratefully acknowledge the use of Matlab code by Markus Svens{\'e}n (to train GTM) and Ian Nabney and Christopher Bishop (Netlab, to train MLPs), both freely available at \url{http://www.ncrg.aston.ac.uk}.}
(fig.~\ref{f:exp-toy:results}B):
\begin{itemize}
\item A factor analyser with one factor. We use this as a linear-Gaussian density model baseline.
\item A multilayer perceptron (MLP) with a single hidden layer of $48$ units, trained to minimise the squared reconstruction error using stochastic gradient descent and small, random starting values for the weights. We use this as universal mapping approximator baseline.
\item A 1D GTM with a grid of $K = 200$ points and $9$ Gaussian basis functions of width equal to the separation between basis functions centres, all in the $[-1,1]$ interval (see \citealp{Bishop_98a}). This results in a Gaussian mixture with $K = 200$ components all with the same, isotropic covariance. We use this to implement mean- and mode-based methods.
\end{itemize}
We compared the following reconstruction methods based on GTM:
\begin{itemize}
\item[\texttt{mean}] Single pointwise reconstruction by the conditional mean (section~\ref{s:deriv-funct-rel:mean}).
\item[\texttt{gmode}] Single pointwise reconstruction by the global mode of the conditional distribution.
\item[\texttt{rmode}] Single pointwise reconstruction by a random mode (all modes are taken equally likely).
\item[\texttt{cmode}] Single pointwise reconstruction by the \emph{closest mode}, i.e., the mode of the conditional distribution that is closest in Euclidean distance to the true value of the original sequence (of course, unknown in practice). The \texttt{cmode} gives a lower bound of the reconstruction error achievable by any mode-based method (\texttt{gmode}, \texttt{rmode}, \texttt{grmode}, \texttt{dpmode}) and tells us how much usable reconstruction information is contained in the conditional modes.
\item[\texttt{grmode}] Single pointwise reconstruction by the mode of the conditional distribution that is closest in Euclidean distance to the previously reconstructed point, i.e., a greedy minimisation of $\mathscr{C}$ (section~\ref{s:constr:min-greedy}). 
\item[\texttt{dpmode}] Multiple pointwise reconstruction by the modes of the conditional distribution followed by dynamic programming minimisation of $\mathscr{C}$ to select the global reconstruction (section~\ref{s:constr:min-dp}). The continuity constraint $\mathscr{C}$ is based on the unweighted Euclidean distance, eq.~\eqref{e:constr:min:cont}. This is the method we advocate in section~\ref{s:method}.
\item[\texttt{meandp}] Like \texttt{dpmode} except that if the conditional distribution is unimodal, we use its mean rather than its mode. This is intended to account for skewed unimodal conditional distributions.
\item[\texttt{sampdp}] Multiple pointwise reconstruction by $S = 6$ samples of the conditional distribution (section~\ref{s:deriv-funct-rel:sampling}) followed by dynamic programming minimisation of $\mathscr{C}$ to select the global reconstruction. We took $S$ slightly larger than the maximal number of inverse branches of $\g^{-1}$ (equal to $3$) so that all the branches have a chance to contribute but without facilitating the appearance of outliers.
\end{itemize}
Additionally, we compared with the methods \texttt{fa} for factor analysis, for which the mean- and mode-based methods coincide (being a symmetric, unimodal density); and \texttt{mlp} for the multilayer perceptron (but only for masks $\M_{\text{fwd}}$ and $\M_{\text{inv}}$, which correspond to regression patterns of missing data).

\begin{FPfigure}
    \psfrag{t1}[][]{$t_1$}
    \psfrag{t2}[t][t]{$t_2$}
    \psfrag{TRset}[rB][rB]{Training set}
    \psfrag{noisytraj}[rB][rB]{\caja{c}{c}{Noisy \\ trajectory}}
    \psfrag{mapping}[tl][tl]{\caja{c}{c}{Mapping \\ $t_2 = g(t_1)$}}
    \psfrag{GTMdensity}[rB][rB]{\caja{t}{c}{GTM density \& \\ latent manifold}}
    \psfrag{FAdensity}[lB][lB]{\caja{t}{c}{FA density \& \\ latent manifold}}
    \psfrag{MODE}[B][lB]{\footnotesize mode}
    \psfrag{MEAN}[][]{\footnotesize mean}
    \psfrag{GTMcond}{\caja{c}{c}{cond.~distr. \\ $p(t_1|t_2)$, GTM}}
    \psfrag{FAcond}{\caja{c}{c}{cond.~distr. \\  $p(t_1|t_2)$, FA}}
    \psfrag{Minv}[][]{Mask $\M_{\text{inv}}$}
    \psfrag{M50}[][]{Mask $\M_{\text{$50$\%}}$}
    \psfrag{Mfwd}[][]{Mask $\M_{\text{fwd}}$}
    \psfrag{original}{original}
    \psfrag{mean}{\texttt{mean}}
    \psfrag{fa}{\texttt{fa}}
    \psfrag{mlp}{\texttt{mlp}}
    \psfrag{gmode}{\texttt{gmode}}
    \psfrag{grmode}{\texttt{grmode}}
    \psfrag{dpmode}{\texttt{dpmode}}
    \psfrag{cmode}{\texttt{cmode}}
    \psfrag{sampdp}{\texttt{sampdp}}
    \psfrag{conddist}[B][B]{Cond.~distr.\ $p(t_2|t_1)$}
    \psfrag{modes}[rB][rB]{modes}
    \psfrag{x-2pi}[][Bc]{\raisebox{-0.15cm}[0pt][0pt]{\tiny $-2\pi$}}
    \psfrag{x-4pi3}[][Bc]{\raisebox{-0.15cm}[0pt][0pt]{\tiny $-\frac43\pi$}}
    \psfrag{x-2pi3}[][Bc]{\raisebox{-0.15cm}[0pt][0pt]{\tiny $-\frac23\pi$}}
    \psfrag{x0pi}[][Bc]{\raisebox{-0.15cm}[0pt][0pt]{\tiny $0$}}
    \psfrag{x2pi3}[][Bc]{\raisebox{-0.15cm}[0pt][0pt]{\tiny $\frac23\pi$}}
    \psfrag{x4pi3}[][Bc]{\raisebox{-0.15cm}[0pt][0pt]{\tiny $\frac43\pi$}}
    \psfrag{x2pi}[][Bc]{\raisebox{-0.15cm}[0pt][0pt]{\tiny $2\pi$}}
    \psfrag{y-2pi}[r][r]{\tiny $-2\pi$\hspace*{0.1cm}}
    \psfrag{y-4pi3}[r][r]{\tiny $-\frac43\pi$\hspace*{0.1cm}}
    \psfrag{y-2pi3}[r][r]{\tiny $-\frac23\pi$\hspace*{0.1cm}}
    \psfrag{y0pi}[r][r]{\tiny $0$\hspace*{0.1cm}}
    \psfrag{y2pi3}[r][r]{\tiny $\frac23\pi$\hspace*{0.1cm}}
    \psfrag{y4pi3}[r][r]{\tiny $\frac43\pi$\hspace*{0.1cm}}
    \psfrag{y2pi}[r][r]{\tiny $2\pi$\hspace*{0.1cm}}
    \psfrag{AA}{\textbf{A}}
    \psfrag{BB}{\textbf{B}}
    \psfrag{CC}{\textbf{C}}
    \psfrag{DD}{\textbf{D}}
    \psfrag{EE}{\textbf{E}}
    \psfrag{FF}{\textbf{F}}
    \psfrag{GG}{\textbf{G}}
    \psfrag{HH}{\textbf{H}}
    \psfrag{II}{\textbf{I}}
    \includegraphics[width=\textwidth]{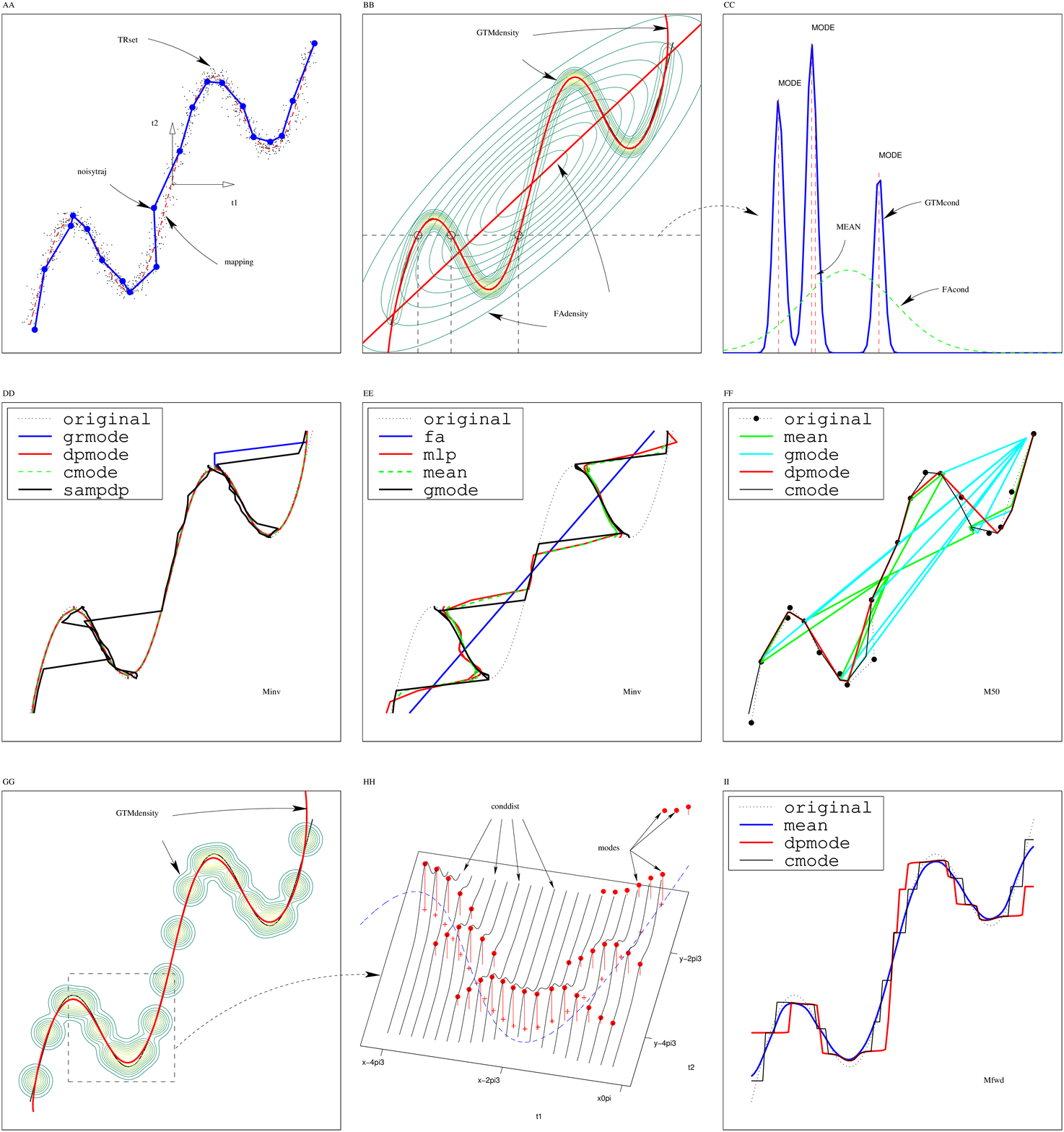}
    \caption{Reconstruction results for the toy problem of section~\ref{s:exp-toy}. This figure is best viewed in colour. All panels (except \textbf{C} and \textbf{H}) show the rectangle $(t_1,t_2) \in [-2\pi,2\pi] \times [-2\pi,2\pi]$. The panels are as follows (see main text for details). \textbf{A}: the training set for the density models (dots), the underlying data manifold (dashed line) and a sample noisy trajectory (solid circles). \textbf{B}: contour plots of the joint density models: GTM (with $K = 200$ Gaussian components) and FA. \textbf{C}: the conditional distributions $p(t_1|t_2)$ for a value $t_2 = -3.8$ (horizontal dashed line in panel \textbf{B}). For GTM, this conditional distribution contains $3$ modes (corresponding to the circles in \textbf{B}) while for FA it is a broad Gaussian whose mean falls out of the data. \textbf{D, E}: reconstruction of a noiseless trajectory with $N = 100$ points (original) by several methods, for the mask $\M_{\text{inv}}$ (i.e., for each point in the trajectory, $t_2$ is present and $t_1$ is missing). In \textbf{D}, the original trajectory is almost indistinguishable from \texttt{dpmode} and \texttt{cmode}, while \texttt{grmode} and \texttt{sampdp} produce reconstructed trajectories with retracings and shortcuts. In \textbf{E}, the methods fail: \texttt{fa} returns a linear trajectory, while \texttt{gmode}, \texttt{mlp} and \texttt{mean} return a univalued mapping with discontinuous jumps between branches (see the figure sideways). \textbf{F}: reconstruction of the noisy trajectory of panel \textbf{A} with $N = 20$ points (original) by several methods, for the mask $\M_{\text{$50$\%}}$ (i.e., for each point in the trajectory, any of $t_1$ or $t_2$ is missing $50$\% of the times). \texttt{cmode} and \texttt{dpmode} give very good reconstructions, while \texttt{mean} and \texttt{gmode} fail. The jumps to the point in the centre (\texttt{mean}) and the top-right corner (\texttt{gmode}) occur when both $t_1$ and $t_2$ are missing at one point of the trajectory. \textbf{G}: GTM density model with only $K = 20$ Gaussian components, resulting in a nonsmooth density. \textbf{H}: blowup of the box in panel \textbf{G} showing the conditional distributions $p(t_2|t_1)$ at several values of $t_1$ (the red lines) and their modes (the circles $\bullet$) and means (the crosses $+$). The dotted line is the data manifold. Note how where the Gaussian components are widely separated, $p(t_2|t_1)$ has more than one mode even though $t_1 \rightarrow t_2$ is univalued. \textbf{I}: reconstruction results for the noiseless trajectory using the nonsmooth GTM model of panel \textbf{G}, for several methods and mask $\M_{\text{fwd}}$ (i.e., for each point in the trajectory, $t_1$ is present and $t_2$ is missing).}
    \label{f:exp-toy:results}
\end{FPfigure}

We ran a number of experiments of which we discuss a representative selection (fig.~\ref{f:exp-toy:results}, table~\ref{t:error}). The same basic results were confirmed with many randomly generated training sets, masks and trajectories to be reconstructed. Additional experiments are given in \citet{Carreir01a}, including further masks (e.g.\ data missing by blocks) and models (e.g.\ full-covariance Gaussian mixtures, MLP ensembles). We report various aspects of the results next.

\paragraph{Method comparison} Based on these experiments we can draw the following conclusions about the methods:
\begin{itemize}
\item As expected, \texttt{fa} is always much worse than \texttt{mean} for any mask, since the forward mapping is nonlinear; and for both masks $\M_{\text{fwd}}$ and $\M_{\text{inv}}$, \texttt{mlp} was practically equal to the \texttt{mean}.
\item Since our GTM model is very close to the true density, the \texttt{mean} approximates extremely well the forward mapping (mask $\M_{\text{fwd}}$, not shown in fig.~\ref{f:exp-toy:results}), being univalued. It fails with the inverse mapping (mask $\M_{\text{inv}}$, fig.~\ref{f:exp-toy:results}E), this being multivalued: the univalued \texttt{mean} mapping travels through the midpoints of the inverse branches, blending them into a single branch. Because of the symmetry of the forward mapping, the midpoint of these branches always happens to coincide with one of the branches and so the result is better than it should be in a general case lacking symmetry (where the \texttt{mean} will not be a valid inverse). The \texttt{mean} also fails for general masks (e.g.\ mask $\M_{\text{$50$\%}}$, fig.~\ref{f:exp-toy:results}F), although, as predicted by the theory, in terms of average reconstruction error it is still the best method based on \emph{single} pointwise reconstruction (the others being \texttt{gmode} and \texttt{rmode}).
\item Both \texttt{gmode} and \texttt{rmode} result in discontinuous mappings, with frequent branch switches, but unlike the \texttt{mean} they always provide with valid inverses, because they track branches. \texttt{gmode} generally outperforms \texttt{rmode}---the latter can be considered as the chance baseline for single pointwise reconstruction by the modes.
\item \texttt{cmode} achieves practically zero reconstruction error for all masks considered, vastly outperforming the \texttt{mean} too (except, marginally, sometimes with mask $\M_{\text{fwd}}$, where \texttt{mean} is optimal). This demonstrates that the modes of a good conditional distribution contain information that can potentially achieve near-zero reconstruction error; the problem lies in the selection of a good constraint that discards the wrong modes.
\item Much of the modes' information is recovered by \texttt{dpmode}, which outperforms or equals any other method, including \texttt{mean}, for any mask. Even for the forward mapping, where \texttt{mean} is guaranteed to be optimal on the average, \texttt{dpmode} still performs as well as the \texttt{mean} (it actually outperforms it in table~\ref{t:error}A, row $\M_{\text{fwd}}$, but this is an isolated instance). Its performance is degraded only slightly for very high amounts of missing data (e.g.\ mask $\M_{\text{$75$\%}}$), where the other methods incur huge errors. For regression problems (masks $\M_{\text{fwd}}$ and $\M_{\text{inv}}$), \texttt{dpmode} may perform worse than \texttt{mean} in two situations analysed below: nonsmooth density models and oversampled trajectories.
\item There is virtually no difference between \texttt{meandp} and \texttt{dpmode}. This is due to the fact that the training set contained isotropic noise, so that when the conditional distribution is unimodal, it is approximately symmetric and its mean and mode nearly coincide. For more complex types of noise \texttt{meandp} may slightly improve over \texttt{dpmode}.
\item Both \texttt{grmode} and \texttt{sampdp} result most times in wrongly reconstructed trajectories that retrace themselves and contain shortcuts between branches. For \texttt{grmode} the reason is the inability to backtrack out of a wrong solution, although for general missing data patterns ($\M_{\text{$75$\%}}$--$\M_{\text{$25$\%}}$) its performance is not much worse than that of \texttt{dpmode}. For \texttt{sampdp} there are two reasons: the inability to find a priori a good value%
\footnote{To force all mapping branches to be represented, we also tried a very high value $S = 100$. The resulting trajectories were smoother but still wrong.}
for the number of samples $S$, so that suboptimal candidate reconstructions are generated and/or correct ones are missed; and the appearance of wrong trajectory reconstructions with low value for the global constraint. Therefore, despite the computational economy of these approaches, they are not recommended.
\end{itemize}

\paragraph{Denoising}

A noisy trajectory is reconstructed as a smooth trajectory because by reducing a conditional distribution to a point (single pointwise reconstruction) or a point per branch (multiple pointwise reconstruction) all variation is eliminated for the given values of the present variables. In fact, a large part of the reconstruction error in table~\ref{t:error} is due to the noise in the original trajectory, which has been removed from the reconstructed one.

\paragraph{Regression is harder than varying patterns of missing data}

For methods based on global constraint minimisation, in particular \texttt{dpmode}, a varying missing data pattern helps to break the ambiguity. The reason is the changing structure of the candidate reconstructions for varying patterns of missing data. When the pattern of missing data is constant (regression-type) and the conditional distribution has spurious modes, it is possible to have long runs of wrong candidate reconstructions that give a short trajectory segment that is shorter than the correct one (even though there may be long jumps where the conditional distribution becomes unimodal). For varying patterns of missing data the spatial structure of these series typically changes dramatically from $n$ to $n+1$. Thus, the runs of wrongly reconstructed points are much shorter and when concatenated they give a longer trajectory than the correct one. This can be seen in table~\ref{t:error} for the \texttt{dpmode}: large errors appear only for nonsmooth density models (table~\ref{t:error}B; see below) or oversampling for the regression-type patterns (masks $\M_{\text{fwd}}$, $\M_{\text{inv}}$), but never for general ones ($\M_{\text{$75$\%}}$--$\M_{\text{$25$\%}}$), even when as many as $76$\% of the values are missing. Thus, the \texttt{dpmode} method is very robust for varying patterns of missing data even with not very good density models, oversampling or large amounts of missing data.

\begin{table}[t]
  \begin{center}
    \begin{tabular}{@{}|l||r|r|rrrrrrrr|@{}}
\multicolumn{11}{l}{\textbf{A}: toy problem} \\
\hline
Mask & \multicolumn{1}{c|}{\texttt{fa}} & \multicolumn{1}{c|}{\texttt{mlp}} & \multicolumn{8}{c|}{GTM with $K=200$ components} \\
\cline{4-11}
 & & & \multicolumn{1}{c}{\texttt{mean}} & \multicolumn{1}{c}{\texttt{gmode}} & \multicolumn{1}{c}{\texttt{rmode}} & \multicolumn{1}{c}{\texttt{cmode}} & \multicolumn{1}{c}{\texttt{grmode}} & \multicolumn{1}{c}{\texttt{dpmode}} & \multicolumn{1}{c}{\texttt{sampdp}} & \multicolumn{1}{c|}{\texttt{meandp}} \\
\hline\hline
$\M_{\text{fwd}}$           &  3.8011 &  0.0129 &  0.0196 &  0.0120 &  0.0120 &  0.0120 &  0.0120 &  0.0120 &  0.2027 &  0.0196 \\
$\M_{\text{inv}}$           &  4.2702 &  2.1633 &  2.1184 &  2.0878 &  7.7086 &  0.0129 &  0.7529 &  0.0129 &  2.0003 &  0.0129 \\
$\M_{\text{$75$\%}}$           & 15.5385 &         & 14.6874 & 62.8203 & 36.1892 &  0.0069 &  0.2534 &  0.1936 &  5.9703 &  0.2059 \\
$\M_{\text{$50$\%}}$           & 11.4116 &         &  9.7848 & 31.4508 & 14.8545 &  0.0058 &  0.1512 &  0.0746 &  4.6827 &  0.0867 \\
$\M_{\text{$25$\%}}$           &  2.9049 &         &  1.7891 &  6.5224 &  3.1629 &  0.0040 &  0.0191 &  0.0066 &  0.6252 &  0.0062 \\
\hline
\multicolumn{11}{c}{} \\[.3cm]
\multicolumn{11}{l}{\textbf{B}: toy problem} \\
\hline
Mask & \multicolumn{1}{c|}{} & \multicolumn{1}{c|}{} & \multicolumn{8}{c|}{GTM with $K=20$ components} \\
\cline{4-11}
 & & & \multicolumn{1}{c}{\texttt{mean}} & \multicolumn{1}{c}{\texttt{gmode}} & \multicolumn{1}{c}{\texttt{rmode}} & \multicolumn{1}{c}{\texttt{cmode}} & \multicolumn{1}{c}{\texttt{grmode}} & \multicolumn{1}{c}{\texttt{dpmode}} & \multicolumn{1}{c}{\texttt{sampdp}} & \multicolumn{1}{c|}{\texttt{meandp}} \\
\hline\hline
$\M_{\text{fwd}}$           & & &  0.0956 &  0.1659 &  2.0207 &  0.1609 &  2.1708 &  1.1226 &  0.3732 &  1.1338 \\
$\M_{\text{inv}}$           & & &  2.2376 &  3.3735 &  8.1796 &  0.1048 &  3.8005 &  0.1093 &  2.1427 &  0.1094 \\
$\M_{\text{$75$\%}}$           & & & 14.8596 & 58.0260 & 31.1647 &  0.1566 &  1.0583 &  0.2272 &  8.5859 &  0.2285 \\
$\M_{\text{$50$\%}}$           & & &  9.7731 & 28.7667 & 21.0178 &  0.1168 &  0.5550 &  0.1181 &  2.5967 &  0.1198 \\
$\M_{\text{$25$\%}}$           & & &  1.8337 &  6.1947 &  7.9984 &  0.0510 &  0.0984 &  0.0510 &  0.6293 &  0.0517 \\
\hline
\multicolumn{11}{c}{} \\[.3cm]
\multicolumn{11}{l}{\textbf{C}: robot arm problem} \\
\hline
Mask & \multicolumn{1}{c|}{\texttt{fa}} & \multicolumn{1}{c|}{\texttt{mlp}} & \multicolumn{8}{c|}{GTM with $K=225$ components} \\
\cline{4-11}
 & & & \multicolumn{1}{c}{\texttt{mean}} & \multicolumn{1}{c}{\texttt{gmode}} & \multicolumn{1}{c}{\texttt{rmode}} & \multicolumn{1}{c}{\texttt{cmode}} & \multicolumn{1}{c}{\texttt{grmode}} & \multicolumn{1}{c}{\texttt{dpmode}} & \multicolumn{1}{c}{\texttt{sampdp}} & \multicolumn{1}{c|}{\texttt{meandp}} \\
\hline\hline
$\M_{\text{fwd}}$           & 0.0130 & 0.0007 & 0.0014 & 0.0017 & 0.0017 & 0.0017 & 0.0017 & 0.0017 & 0.0027 & 0.0014 \\
$\M_{\text{inv}}$           & 0.7690 & 0.6394 & 0.6767 & 1.2998 & 1.4603 & 0.0057 & 3.4837 & 0.3230 & 0.4602 & 0.3230 \\
$\M_{\text{$75$\%}}$           & 0.7369 &        & 0.7084 & 1.9592 & 1.7930 & 0.0072 & 0.2395 & 0.0928 & 0.1366 & 0.0928 \\
$\M_{\text{$50$\%}}$           & 0.4136 &        & 0.3655 & 1.1159 & 1.1873 & 0.0045 & 0.4156 & 0.1092 & 0.0717 & 0.1094 \\
$\M_{\text{$25$\%}}$           & 0.2297 &        & 0.1486 & 0.1539 & 0.2371 & 0.0029 & 0.0343 & 0.0074 & 0.0147 & 0.0080 \\
\hline
    \end{tabular}
    \caption{Reconstruction results: average squared error $\frac{1}{N}\sum^N_{n=1}{\norm{\t^{(n)}-\hat{\t}^{(n)}}^2}$, where $\{\t^{(n)}\}^N_{n=1}$ is the original trajectory and $\{\hat{\t}^{(n)}\}^N_{n=1}$ the reconstructed one. The results are given for different methods and masks (see the main text for definitions of these). \textbf{A}: toy problem with a smooth density model. \textbf{B}: toy problem with a nonsmooth density model. \textbf{C}: robot arm problem.}
    \label{t:error}
  \end{center}
\end{table}

\paragraph{Nonsmooth density models and spurious modes}

In the previous experiments we have used a nearly ideal density model (GTM with $K = 200$ components): it approximates the true density almost exactly and so any conditional distribution has the right number of modes and at the right locations. Gaussian mixtures, being a superposition of localised bumps, have a tendency to develop ripple on an otherwise smooth density, as seen with a GTM model of $K = 20$ components (fig.~\ref{f:exp-toy:results}G). Although the density estimate is worse than that of fig.~\ref{f:exp-toy:results}B in terms of log-likelihood, it still represents qualitatively well the density. However, the mixture components do not coalesce enough in some regions. This results in wavy conditional distributions having more modes than they should (see $p(t_2|t_1)$ for various values of $t_1$ in fig.~\ref{f:exp-toy:results}H). Some of the true modes along the trajectory have unfolded into a few modes scattered in a small area around the true mode. A very good reconstruction is still possible since some of these modes are very close to the true one, as evidenced by the low reconstruction error of the \texttt{cmode}. The \texttt{mean} also achieves a reconstruction error about as low as with $K = 200$, being largely insensitive to the ripple. But for mask $\M_{\text{fwd}}$ the error for \texttt{dpmode} is now $1.1226$ for $K = 20$ while it was $0.0120$ for $K = 200$ (fig.~\ref{f:exp-toy:results}I). The problem is that this crowd of spurious modes may well allow wrong reconstructed trajectories that have a lower global constraint value (that are shorter) due to \emph{shortcuts} that appear as horizontal and nearly vertical segments in fig.~\ref{f:exp-toy:results}I. The parameter that governs this behaviour is the ratio between the extent of the mode scatter inside a conditional distribution and the sampling period of the trajectory: the larger the scatter, the more likely interference becomes with neighbouring trajectory points. (Owing to the geometry of this particular example, no spurious modes appear in the conditional distribution $t_1|t_2$ and so the error for $\M_{\text{inv}}$ remains low.) The error for general missing data patterns remains low for the same reason as before: the subsets of missing variables usually change from point $n$ to point $n+1$ and thus the probability of getting a run of several points whose conditional distributions have spurious modes decreases. In general, spurious modes can also appear when using Gaussian components with separate, full-covariance parameters (which, besides, are much harder to train due to log-likelihood singularities).

\paragraph{Over- and undersampling}

We experimented with very small and very large values of the sampling rate of the trajectory for method \texttt{dpmode} (or equivalently, for very large and very small values of the number of points $N$ in the trajectory, respectively). A very small sampling rate is one close to the Nyquist rate; a very large sampling rate is one whose period is much smaller than the noise (normal with $\sigma = 0.2$). For undersampling, \texttt{dpmode} still reconstructs well the trajectory up to $N = 20$ but starts finding wrongly reconstructed trajectories for lower $N$, particularly for the worse GTM models ($K = 20$). This is clearly due to a lack of enough information to reconstruct the trajectory. More surprising are the results for oversampling: for $N = 1\,000$, \texttt{dpmode} can give wrongly reconstructed trajectories that retrace themselves and have shortcuts for mask $\M_{\text{inv}}$ (and, for nonsmooth models, for $\M_{\text{fwd}}$ too) although it still reconstructs well for $\M_{\text{$75$\%}}$--$\M_{\text{$25$\%}}$. The reason is that the original trajectory is polygonally very long (the $1\,000$-point trajectory is $12$ times longer than the $100$-point one: high $\mathscr{C}$) but is not long in terms of actual displacement---being a random walk superimposed on a slow drift, it twists around itself many times in a region of size $\sigma$. Thus, if there are multiple pointwise candidate reconstructions, there often exist shorter trajectories containing multiple retracings of a branch segment and infrequent branch switchings. The quality of the density model is not really at fault here: it is a characteristic of the global constraint chosen. We found that the trajectories were correctly reconstructed if we used the squared Euclidean distance in the value of $\mathscr{C}$ (instead of the Euclidean distance), the reason being that long jumps are then penalised more.

\paragraph{Other effects}

The reconstructed trajectories tend to show a slight error at the trajectory turns, e.g.\ in fig.~\ref{f:exp-toy:results}D for \texttt{dpmode} and \texttt{grmode}, or in fig.~\ref{f:exp-toy:results}I for \texttt{mean}. The error consists of cutting short through the turns (for all methods) for mask $\M_{\text{fwd}}$ and, less noticeably, of a spike right at the tip of the turns (for \texttt{grmode} and \texttt{dpmode}) for mask $\M_{\text{inv}}$. The ``cutting-short'' effect is due to slight imperfections of the GTM density estimate. The Gaussian components interact more strongly in the convex side of the turn, pile up there and bias the mean (see fig.~\ref{f:exp-toy:results}H); cutting through turns of the original trajectory also saves trajectory length. The spike is the premature blending of two inverse branches into one branch. As the two branches approach their intersection point, the bumps associated with the two respective modes of the conditional distribution interact and blend into a single unimodal bump before the intersection point. In both cases the effect size is larger the larger the component variance ($\sigma^2$) is; in turn, this variance is larger the noisier the training set is and the fewer components ($K$) are used.

With general missing data patterns, the case of all variables ($t_1$, $t_2$) missing at a point $n$ in the sequence results in two different behaviours. Single pointwise reconstruction methods prescribe reconstructing them with a fixed value: the mean of the joint density model for \texttt{fa} and \texttt{mean} and its global mode for \texttt{gmode}. This produces large jumps to that fixed value and thus inflates the reconstruction error (fig.~\ref{f:exp-toy:results}F). Multiple pointwise reconstruction methods prescribe reconstructing them with all the modes of the joint density model, of which there are $19$ and $6$ for $K = 20$ and $200$, respectively. This adds more flexibility and reduces the reconstruction error, since the jumps are now to one of those modes and are therefore shorter. Even better results are obtained by using all $K$ centroids instead of the modes, particularly for very smooth density models where the components coalesce strongly and decrease the number of modes. Strictly, though, the case of all variables missing is just a particular case, the most extreme one, of a range of missingness patterns.

\paragraph{Summary}

The main result is that, for a good density model and if continuity holds, \texttt{dpmode} (our method) can greatly improve over the traditional methods \texttt{mean} (the conditional mean) and \texttt{mlp} (the universal mapping approximator), approaching the limit of \texttt{cmode} (which is close to zero error) for all patterns of missing data; and is particularly successful for general patterns of missing data even for poor density models, oversampled trajectories or large amounts of missing data. This means that the modes contain important information about the possible options to predict the missing values, and that application of the continuity constraint allows to recover that information.

If the density model is not smooth, the conditional distribution presents spurious modes which may give rise to wrong solutions of the dynamic programming search. In this case, the reconstruction error with \texttt{dpmode} for regression problems ($\M_{\text{fwd}}$, $\M_{\text{inv}}$) usually exceeds that of the conditional mean. For general patterns of missing data ($\M_{\text{$75$\%}}$--$\M_{\text{$25$\%}}$) the error increase is small. The \texttt{mean} is barely affected in any case.

Finally, oversampling seems: (1) not to affect the \texttt{dpmode} for general missing data patterns (for both smooth and nonsmooth density models); (2) to introduce quantisation errors for forward (univalued) mappings but only in some areas, with the overall reconstruction being correct (the smoother the model, the lower the error); and (3) to severely degrade the quality of the reconstruction for inverse multivalued mappings due to shortcuts and retracings (for both smooth and nonsmooth density models).

\section{Experiments: robot arm inverse kinematics}
\label{s:exp-robotarm}

\begin{figure}
  \begin{center}
    \psfrag{l1}[][]{$l_1$}
    \psfrag{l2}[][]{$l_2$}
    \psfrag{x1}[][]{$x_1$}
    \psfrag{x2}[][]{$x_2$}
    \psfrag{t1}[][]{$\theta_1$}
    \psfrag{t2}[][]{$\theta_2$}
    \psfrag{end-effector}[t][b]{\caja{t}{c}{end effector \\ $(x_1,x_2)$}}
    \psfrag{elbow-up}[][]{\caja{t}{c}{``elbow-up'' \\ configuration}}
    \psfrag{elbow-down}[][]{\caja{t}{c}{``elbow-down'' \\ configuration}}
    \begin{tabular}{@{}c@{\hspace*{0.15\textwidth}}c@{}}
    \includegraphics[height=0.3\textwidth]{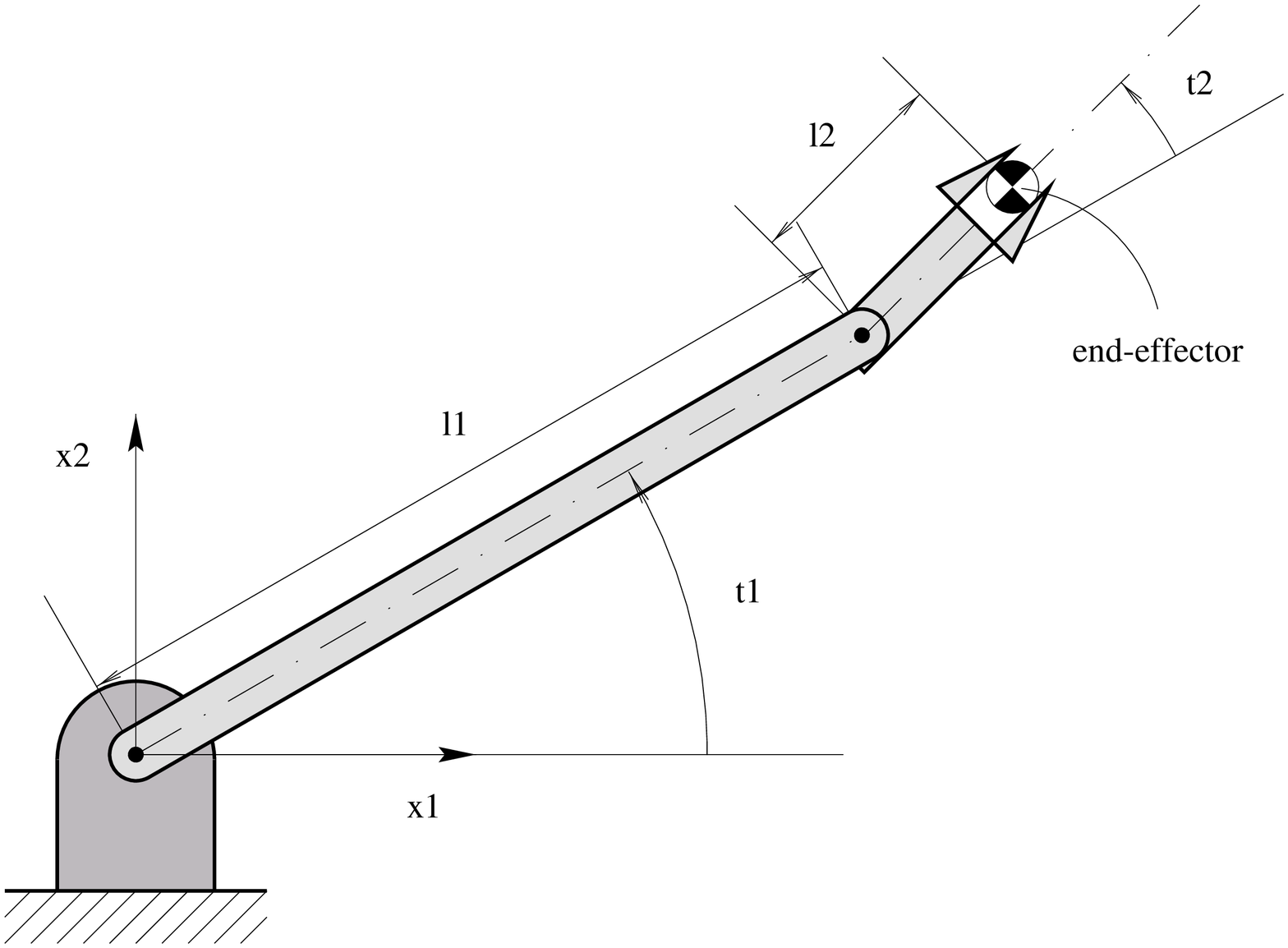} &
    \psfrag{end-effector}[][]{end effector}
    \includegraphics[height=0.3\textwidth]{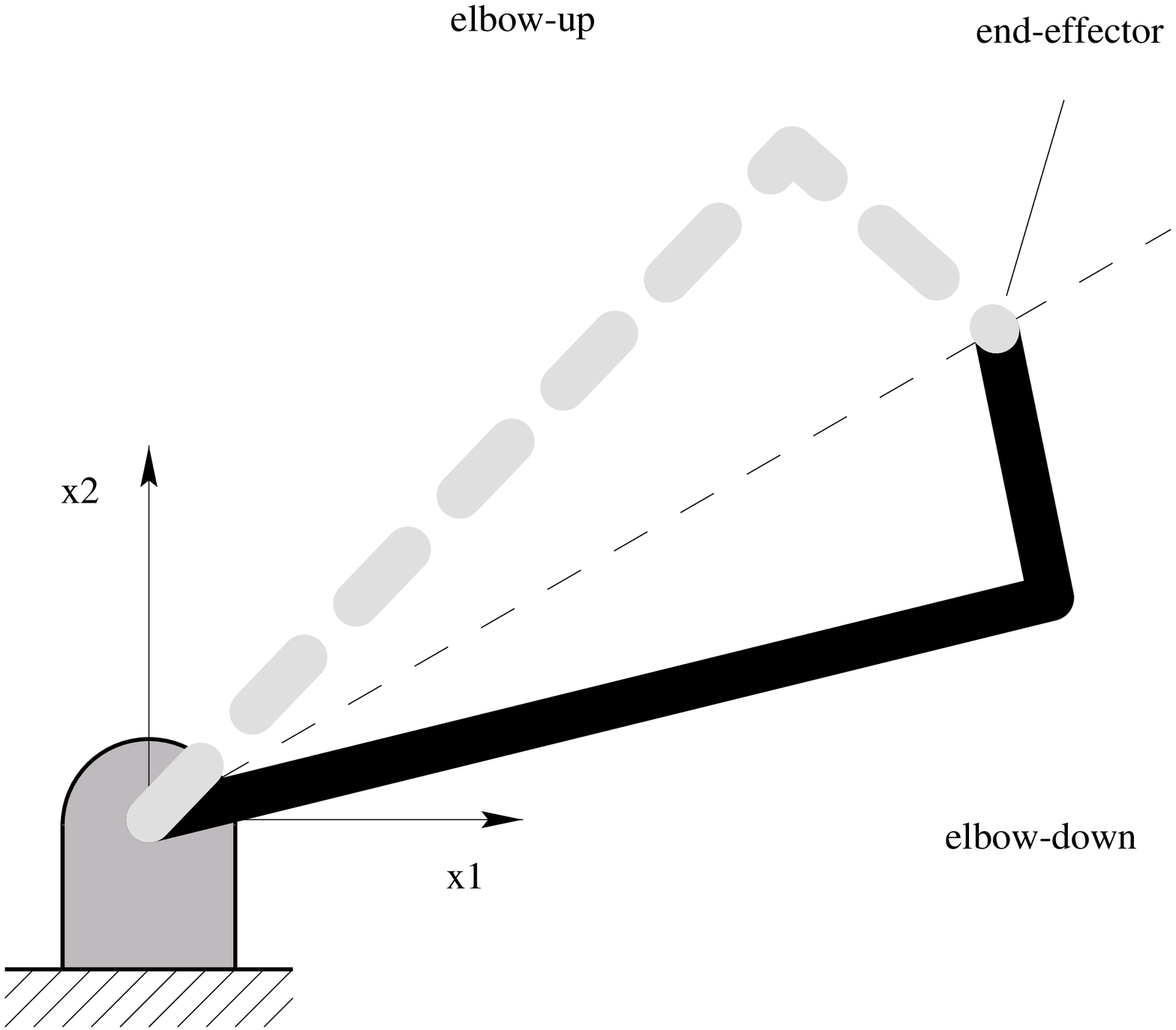}
    \end{tabular}
    \caption{\emph{Left}: schematic of a two-link, planar robot arm of joint angles $(\theta_1, \theta_2)$ and end-effector position $(x_1, x_2)$. \emph{Right}: two different configurations of the joint angles that yield the same end-effector position.}
    \label{f:exp-robotarm}
  \end{center}
\end{figure}

The inverse kinematics of a robot arm is a prototypical example of a mapping inversion problem, with a well-defined forward mapping and a multivalued inverse mapping. We consider a two-joint, planar arm that has been often used in the pattern recognition literature (e.g.\ \citealp{Bishop94a,RohwerRest96}). The forward mapping \g\ gives the position in Cartesian coordinates $\x = \left(\begin{smallmatrix} x_1 \\ x_2 \end{smallmatrix}\right) \in \calC$ of the end effector (the hand of the robot arm) given the angles $\btheta = \left(\begin{smallmatrix} \theta_1 \\ \theta_2 \end{smallmatrix}\right) \in \calA$ at its joints, where $\calA = [0.3,1.2] \times [\frac{\pi}{2},\frac{3\pi}{2}]$ is the \emph{actuator space} and $\calC = \g(\calA)$ the \emph{work space} (i.e., the region reachable by the end effector):
\begin{align*}
  x_1 & = l_1 \cos{\theta_1} + l_2 \cos(\theta_1+\theta_2) \\
  x_2 & = l_1 \sin{\theta_1} + l_2 \sin(\theta_1+\theta_2)
\end{align*}
with constant link lengths ($l_1 = 0.8$, $l_2 = 0.2$); see fig.~\ref{f:exp-robotarm}. The transformation from the desired end-effector position to the corresponding joint angles (inverse kinematics) can be obtained analytically for this simple arm \citep{AsadaSlotin86a} and is in general bivalued (``elbow up'' and ``elbow down'' configurations). Studies of trajectory formation have considered sophisticated cost functions such as energy, torque, jerk or acceleration \citep{Nelson83a}, all expressible in terms of quadratic functions of derivatives. Besides, since the forward mapping is known we could further use an $\mathscr{F}$-constraint (see section~\ref{s:constr:fwd-map}). Although we could favourably use these, we will show that a very simple cost function, continuity $\mathscr{C}$ in the space $(\btheta, \x)$, is enough to recover the trajectory.

\begin{figure}
  \begin{center}
    \psfrag{x1}[][]{\raisebox{-0.25cm}[0pt][0pt]{$x_1$}}
    \psfrag{x2}[][]{\raisebox{0.1cm}[0pt][0pt]{$x_2$}}
    \psfrag{theta1}[][]{\raisebox{-0.25cm}[0pt][0pt]{$\theta_1$}}
    \psfrag{theta2}[][]{\raisebox{0.1cm}[0pt][0pt]{$\theta_2$}}
    \begin{tabular}{@{}c@{\hspace*{0.15\textwidth}}c@{}}
    \includegraphics[height=0.47\textwidth]{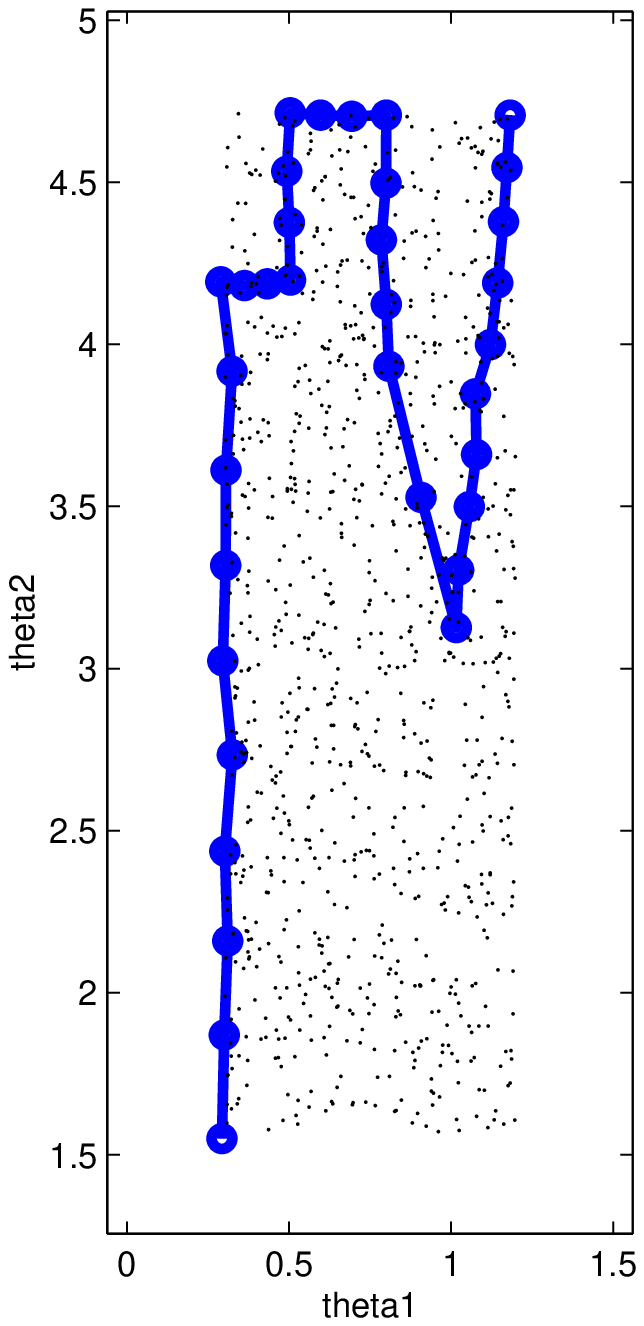} & \includegraphics[width=0.47\textwidth]{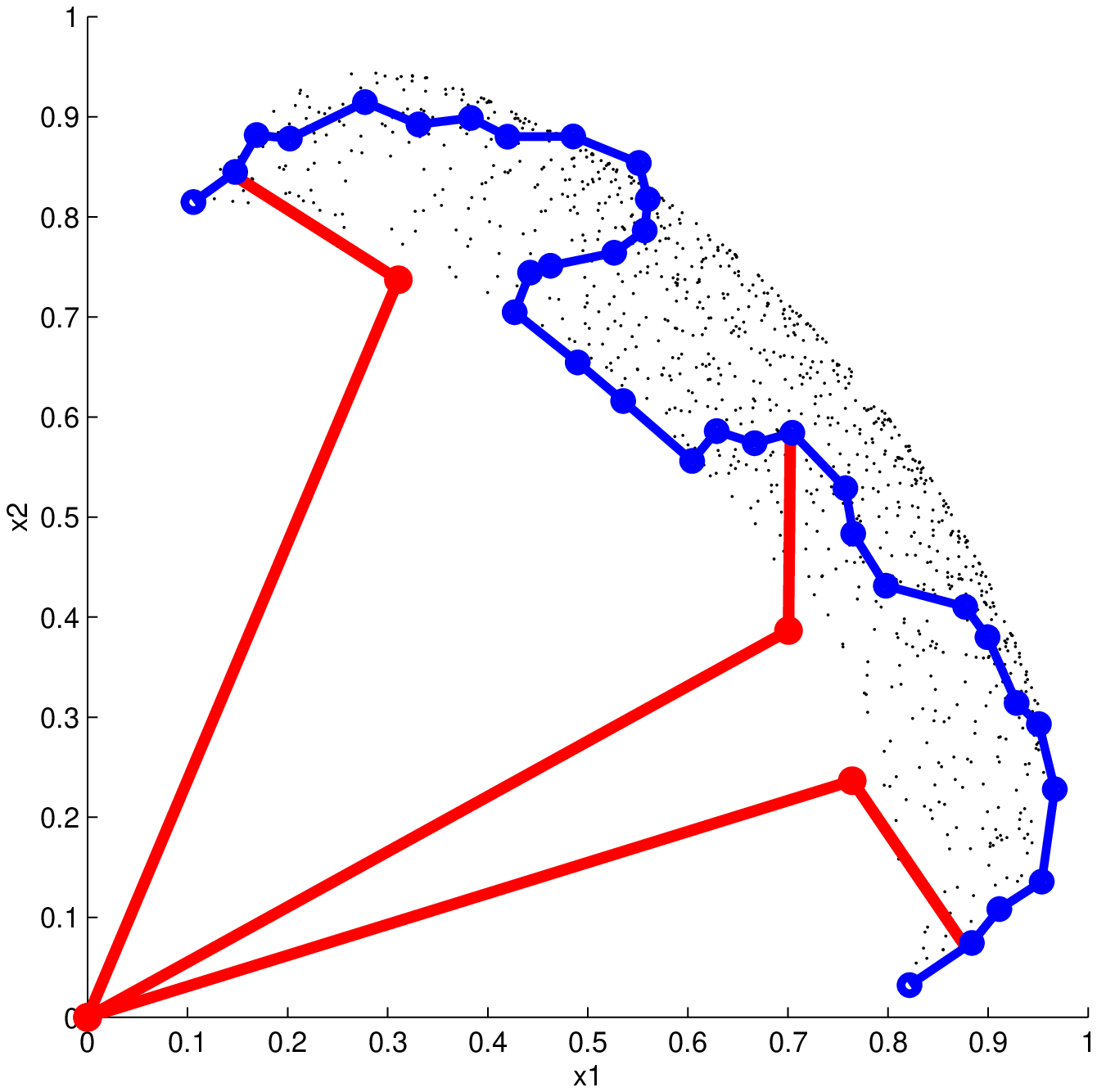}
    \end{tabular}
    \caption{Trajectory of the robot arm end effector to be reconstructed. \emph{Left}: trajectory in actuator space $(\theta_1,\theta_2)$. \emph{Right}: trajectory in work space $(x_1,x_2)$ and three sample robot arm configurations. The training set used is shown in black dots: on the left graph it is a uniform cloud in actuator space; on the right graph it delineates the work space (the region reachable by the end effector).}
    \label{f:exp-robotarm:traj}
  \end{center}
\end{figure}

The experimental methodology is analogous to that of the toy problem. A training set of $N' = 1\,000$ points was generated by sampling uniformly the actuator space, then applying the forward mapping and finally adding normal spherical noise of standard deviation $\sigma = 0.05$ (see fig.~\ref{f:exp-robotarm:traj}). We trained the following models: an MLP with a single hidden layer of $h = 48$ units, a factor analyser with latent space of dimension $L = 2$ and a GTM with latent space of dimension $L = 2$, $15\times 15$ latent grid and $7\times 7$ RBF grid (resulting in a Gaussian mixture with $K = 225$ equal, isotropic components). The number of parameters of the MLP and GTM were similar (around $200$). We applied the same methods and masks as in the toy example, with masks $\M_{\text{fwd}}$ and $\M_{\text{inv}}$ meaning the regressions $\btheta \rightarrow \x$ and $\x \rightarrow \btheta$, respectively. For \texttt{sampdp}, we generated $S = 6$ samples per conditional distribution. We manually designed a continuous trajectory in actuator space (sampled at $N = 34$ points) and obtained the corresponding trajectory in work space by applying \g; we then added small normal noise ($\sigma = 0.01$) to all values to obtain the sequence $\{(\btheta^{(n)},\x^{(n)})\}^N_{n=1}$.

The practically interesting problem (mask $\M_{\text{inv}}$) is to recover $\{\btheta^{(n)}\}^N_{n=1}$ from $\{\x^{(n)}\}^N_{n=1}$ so that impossible movements of the arm (discontinuities in \btheta) are avoided. The reconstruction results, given in table~\ref{t:error}C, show a similar behaviour to that observed in results of the toy experiments: \texttt{dpmode} beats the other methods (in particular, the \texttt{mean} and the \texttt{mlp}, both of which perform very similarly) and its performance is often close to the bound of \texttt{cmode}, even for high amounts of missing data. The largest error for \texttt{dpmode} occurs for the inverse mapping, which confirms that regression problems, when multivalued, are harder than general missing data patterns. All methods perform equally well in the univalued, forward mapping ($\M_{\text{fwd}}$). We observed that $p(\btheta|\x)$ had $2$ to $17$ modes, which means that the density model is not smooth, although in this case it does not seem to affect the \texttt{dpmode}.

The large error of \texttt{gmode} is mainly due to choosing the wrong branch in parts of the trajectory and having discontinuous jumps when joining the correct one. Note that the error reported by e.g.\ \citet{Bishop94a} (who used a \texttt{gmode}-type method) is $\norm{\smash{\x_n - \g(\hat{\btheta}_n)}}$. This error is low when the reconstructed $\hat{\btheta}_n$ maps well to the given $\x_n$ (i.e., $\hat{\btheta}_n$ is a valid inverse of $\x_n$), but disregards discontinuities of the trajectory, which are given by high $\norm{\smash{\btheta_n - \hat{\btheta}_n}}$ or $\norm{\smash{\hat{\btheta}_{n+1} - \hat{\btheta}_n}}$ (i.e., $\hat{\btheta}_n$ may not be the \emph{right} inverse to use at this point $n$).

These results demonstrate that, in this problem, the continuity constraint $\mathscr{C}$ alone can allow good reconstruction with \texttt{dpmode}. However, since the forward mapping is known, an $\mathscr{F}$-constraint can perfectly be incorporated to improve the reconstruction quality. Further advantages of our method in the inverse kinematics problem include the fact that it can encode multiple (unlimited) learned trajectories; that the trajectory length (number of states) is unlimited; and that the trajectory can have any topology. This makes the method useful for localisation or path planning.

\section{Discussion}
\label{s:discussion}

\subsection{Distributions over trajectories}
\label{s:discussion:prob}

Our algorithm operates in two decoupled stages: first generate set of local candidates, then solve a combinatorial optimisation problem to find the global reconstruction. One way to merge both concepts is to define a grand density over the whole sequence, $p_{\text{G}}(\t^{(1)},\dots,\t^{(N)})$, in a constructive way: first generate $\t^{(1)}$ from the joint density $p(\t)$ of sec.~\ref{s:deriv-funct-rel:density}; then, generate $\t^{(2)}$ subject to being in the data manifold, $p(\t)$, but near $\t^{(1)}$, $p_{\mathscr{C}}(\t|\t^{(1)})$. The latter is simply a Gaussian centred in $\t^{(1)}$ with a covariance (say) $\sigma^2\I$, where $\sigma$ would be related to the speed at which the curve $\t = \mathfrak{F}(\z)$ is traversed, and represents the continuity constraint $\mathscr{C}$. And so on for $\t^{(3)},\dots,\t^{(N)}$. In general, we generate a sequence from $\prod^{N}_{n=1}{p(\t^{(n)})} \prod^{N-1}_{n=1}{p_{\mathscr{C}}(\t^{(n+1)}|\t^{(n)})}$ (normalised). This results in a random walk (the term  $p_{\mathscr{C}}(\t^{(n+1)}|\t^{(n)})$) constrained to the data manifold (the term $p(\t^{(n+1)})$). The distance of the continuity constraint (i.e., the covariance matrix of $p_{\mathscr{C}}$) determines the ``sampling period'' of the sequence. We may now attack the problem of global reconstruction directly, by choosing a representative point of $p_{\text{G}}$ (in the sense of section~\ref{s:deriv-funct-rel}) which will give us a likely reconstruction of the trajectory.

If we maximise the logarithm of the grand density $p_{\text{G}}$ over the missing variables $\{\t^{(n)}_{\calM^{(n)}}\}^N_{n=1}$, we find an objective function
\begin{equation}
  \label{e:discussion:prob}
  \sum^N_{n=1}{\ln{p(\t^{(n)})}} - \frac{1}{2\sigma^2} \sum^{N-1}_{n=1}{\norm{\t^{(n+1)} - \t^{(n)}}^2}
\end{equation}
which has the standard form of a fitness term (on the left) and a constraint term (on the right) with weight $1/2\sigma^2$, as e.g.\ in the elastic net \citep{Durbin_89a}, or its generalisation to more sophisticated constraint priors \citep{CarreirGoodhil03a}. Note that we do not maximise over the parameters of the model $p_{\text{G}}$, but over the missing variables; in particular, this means there are no singularities because the objective function is bounded above. We can obtain the method we have proposed in this paper as a limit case of eq.~\eqref{e:discussion:prob}: if $p(\t^{(n)})$ is taken as a sum of deltas centred at the modes of $p\left(\t^{(n)}_{\calM^{(n)}}|\t^{(n)}_{\calP^{(n)}}\right)$, then the search space is restricted to those modes and operates only on the right side term (our continuity constraint).

An objective function over trajectories opens the door for multiple global reconstruction as defined in section~\ref{s:def}. Further, we might think that this grand distribution could be unimodal since there should be less ambiguity in the global reconstruction than in the local ones. If so we could take the mean and be free from local-maxima problems. But the truth is that $p_{\text{G}}$ may have many local maxima where point $\t^{(n)}$ tends to a mode of $p\left(\t^{(n)}_{\calM^{(n)}}|\t^{(n)}_{\calP^{(n)}}\right)$ for all $n$; this will certainly be the case if such conditional distributions are sharply peaked or $\sigma$ is large, and computing its mean would be difficult. Perhaps an optimisation based on annealing $\sigma$ would be helpful here.

Another characteristic of this method is that the candidate pointwise reconstructions (the modes) are weighted by their respective density values, while in our method all modes are equally weighted. This may bias the reconstruction towards highly likely pointwise reconstructions at the expense of the continuity constraint. Finally, we are also left with the choice of the tradeoff parameter $\sigma$. An implementation and evaluation of this method is left for future research.

\subsection{Computational complexity}

Reconstructing a data set of $N$ vectors requires two separate computations: (1) implicitly constructing the layered graph of fig.~\ref{f:constr:min}, i.e., computing the multiple pointwise reconstructions of each point; (2) finding the shortest path on the graph (we do not consider the cost of estimating the joint density model, since this is done offline with a training set and the resulting density can be reused for reconstructing many different data sets). With dynamic programming, the complexity is given by the total number of links in the graph, which for an average case where every layer has $\overline{\nu} \ge 1$ nodes is $\calO(N \overline{\nu}^2)$. This is very fast and is always dominated by the mode finding. Analysing the complexity of the latter is difficult (see \citealp{Carreir01a,Carreir00b} for details). In general, and as confirmed experimentally, a crucial factor is the amount of missing data, since this directly affects the number of modes found at each point $n$ of the sequence.
It is possible to accelerate the mode search by discarding low-probability mixture components in the conditional distribution (see \citealp{Carreir00b,Carreir01a}; we obtained up to $10\times$ speedup with up to $17$\% increase in reconstruction error); or by reducing the number of mixture components at the potential cost of a less accurate density model.

\subsection{Choice of density model: robustness and smoothness}
\label{s:discussion:density-model}

The modes are a key aspect of our approach. However, the mode is not a robust statistic of a distribution since small variations in the distribution shape can have large effects on the location and number of modes. This is related to the \emph{smoothness} of the density model mentioned in section~\ref{s:exp-toy}: with finite mixtures of localised functions, spurious modes can appear as ripple superimposed on a smoothly-varying function. These spurious modes can happen in regions where the mixture components are sparsely distributed and have little interaction; and their probability value can be as high as that of true modes, which rules out the use of a rejection threshold to filter the spurious ones out. Such spurious modes can lead the dynamic programming search to a wrong trajectory with a large reconstruction error, although we observed this only in regression problems, not in general missing data patterns. For regression problems, specially mapping inversion, applying a forward mapping constraint $\mathscr{F}$ (where the forward mapping may be known exactly or derived by a mapping approximator) should prevent spurious modes from forming part of the reconstructed sequence because spurious modes, by definition, will give a high value for the constraint $\mathscr{F}$.

To prevent spurious modes from entering the constraint minimisation at all, a possible solution is to smooth the density model, either globally (at estimation time) or locally (at mode-finding time, i.e., to smooth the conditional distribution). If the density is a Gaussian mixture with spherical components, then smoothing it by convolution with a Gaussian kernel of width $h$ is equivalent simply to adding $h$ to each component width. Globally smoothing can be done at a higher computational cost by increasing the number of components (in GTM the mixture centroids are not trainable parameters and so we incur no loss of generalisation). Alternatively, one can regularise the density by adding a term to the log-likelihood to penalise low variance parameters. In all these cases, selecting how much to smooth so that important modes are not removed is crucial.

A related problem is that of obtaining a Gaussian mixture that represents well the data with a small number of components, for which many methods exist \citep{FigueirJain02a}. However, it is likely that, in their efforts to reduce the number of components, these methods will result in nonsmooth models.

Note that the use of Gaussian mixtures can be considered optimal with respect to avoiding spurious modes in that convolving an arbitrary function with a Gaussian kernel never (in 1D) or almost never (in higher dimension) results in the creation of new modes, which is not true of any other kernel, as is known in scale-space theory (\citealp{Lindeb94a}; \citealp{CarreirWilliam03b,CarreirWilliam03c}). This, the easy computation of conditional distributions and the availability of algorithms for finding all modes \citep{Carreir00b} make Gaussian mixtures very attractive in the present framework---in spite of the fact that the Gaussian distribution (unlike long-tailed distributions) is not robust to outliers in the training set \citep{Huber81}.

\subsection{Dynamical, sequential and time series modelling}
\label{s:discussion:temporal}

Our approach does not attempt to model the temporal evolution of the system. The joint density model $p(\t)$ is estimated statically. The temporal aspect of the data appears indirectly and a posteriori through the application of the continuity constraints to select a trajectory. In this respect, our approach differs from that of dynamical systems or from models based on Markovian assumptions, such as Kalman filters \citep{Harvey91a} or hidden Markov models \citep{RabinerJuang93}.

The fact that the duration or speed of the trajectory plays no role in our algorithm makes it invariant to time warping. That is, the dynamic programming algorithm depends only on the values of the observed variables but not on the experimental conditions and so is independent of the speed at which the trajectory is traversed. It is also independent of direction, since it can be run forwards (from point $1$ to $N$) or backwards with the same result. Therefore, our reconstruction algorithm does not depend on the particular parametrisation of the trajectory, but just on its geometric shape. This is important in the case of speech: it is well known that hidden Markov models are very sensitive to time warpings, i.e., fast or slow speech styles, where the trajectory in speech feature space is the same but is traversed fast or slowly, respectively. Our reconstruction method should then be robust to time warpings.

A time series prediction can be seen as a reconstruction problem where the data set is $\{\t^{(1)}, \t^{(2)}, \dots, \t^{(N+N')}\}$, $\{\t^{(n)}\}^N_{n=1}$ are present and $\{\t^{(n)}\}^{N+N'}_{n=N+1}$ are missing. However, our method is not useful here: the trivial application of a continuity constraint would lead to $\hat{\t}^{(n)} = \t^{(N)}$ $\forall n > N$, i.e., a constant sequence.

\subsection{Bump-finding rather than mode-finding}
\label{s:discussion:bump}

Besides not being a robust statistic, using a mode as a reconstructed point is not appropriate in general because \emph{locally} the optimal value (in the $L_2$ sense) is the mean. That is, if a function is multivalued it will have several branches; in a neighbourhood around a branch the function becomes univalued and so the mean of that branch would be $L_2$-optimal. This suggests that, when the conditional distribution is multimodal, we should look for \emph{bumps} associated with the correct values and take the means of these bumps as reconstructed values instead of the modes---where by bumps we mean fairly concentrated regions where the density is comparatively high. A possible approach to select the bumps would be to decompose a density $p(\t)$ as a mixture $p(\t) = \sum^K_{k=1}{p(k) p(\t|k)}$. Here, $p(\t|k)$ is the density associated with the $k$th bump, which should be localised in the space of \t\ but can be asymmetrical. If $p(\t)$ is modelled by a mixture of Gaussians (as is our case) then the decomposition could be attained by regrouping Gaussian components with a clustering algorithm. This approach would replace the mode finding procedure with a (probably much faster) grouping and averaging procedure.

\subsection{Reconstruction as a preprocessing step}
\label{s:discussion:recons-preproc}

If the missing data reconstruction is a preprocessing step for some other procedure that ordinarily operates on the complete data, then the whole procedure may be suboptimal but faster than marginalising over the missing variables. For example, in a classification task such as speech recognition, one wants to compute $p(C^{(n)}_i|\t^{(n)})$ where $C^{(n)}_i$ is a phoneme class and $\t^{(n)}$ an acoustic feature vector \citep{RabinerJuang93}. Using a hidden Markov model, such probabilities can be computed for every point $n$ in an utterance and an optimal transcription $C^{(1)},\dots,C^{(N)}$ obtained with the Viterbi algorithm. However, if some features are deemed to be missing (due to the presence of noise, for example), then the correct thing to do is to use $p(C^{(n)}_i|\t^{(n)}_{\calP^{(n)}})$, i.e., to marginalise over the missing variables the joint distribution of what is unknown given what is known:
\begin{equation*}
  p(C^{(n)}_i|\t^{(n)}_{\calP^{(n)}}) = \int{p(C^{(n)}_i,\t^{(n)}_{\calM^{(n)}}|\t^{(n)}_{\calP^{(n)}}) \, d\t^{(n)}_{\calM^{(n)}}} = \int{p(C^{(n)}_i|\t^{(n)}_{\calM^{(n)}}, \t^{(n)}_{\calP^{(n)}}) p(\t^{(n)}_{\calM^{(n)}} | \t^{(n)}_{\calP^{(n)}}) \, d\t^{(n)}_{\calM^{(n)}}}.
\end{equation*}
If we reconstruct the data as $\hat{\t}^{(n)}$ and then use $p(C^{(n)}_i|\hat{\t}^{(n)})$ instead, we are implicitly wasting all the information contained in the distribution $p(\t^{(n)}_{\calM^{(n)}} | \t^{(n)}_{\calP^{(n)}})$, effectively replacing it with a delta function at $\hat{\t}^{(n)}_{\calM^{(n)}}$. \citet{Cooke_01a} have demonstrated empirically the superiority of the marginalisation approach for classification in the context of recognition of occluded speech.

However, strictly what we have shown is that reconstructing and then classifying is worse only when the reconstruction is done on a point-by-point basis, i.e., considering the speech frames independent from each other---which they are not. Thus, there may indeed be a benefit in using a global, utterance-wide constraint to reconstruct the whole speech segment---ideally recovering the original speech---and then classifying it; in other words, reconstructing $\t^{(n)}_{\calM^{(n)}}$ not just from $\t^{(n)}_{\calP^{(n)}}$, but from $\{\t^{(n)}_{\calP^{(n)}}\}^N_{n=1}$, as proposed in our method.

\subsection{Reconstruction via dimensionality reduction}
\label{s:discussion:rec-via-dr}

Continuous latent variable models are a convenient probabilistic formulation of the problem of dimensionality reduction (see \citealp{Carreir01a} for a review). Here, the density in the space of the observed variables \t\ is represented as $p(\t) = \int{p(\t|\x) p(\x) \, d\x}$ where \x\ are the latent variables (with prior distribution $p(\x)$) and $\t|\x$ is a noise model on a low-dimensional manifold defined by a mapping $\t = \f(\x)$ from latent to data space, such as $\t|\x \sim \calN(\f(\x),\bSigma)$. Dimensionality reduction of an observed point \t\ is achieved by taking a representative point $\hat{\x}$ (typically the mean) of the posterior distribution in latent space $p(\x|\t)$.

If a latent variable model is used as density model in our method, one would expect that there is a sequence in latent space that corresponds to the sequence $\{\t^{(n)}\}^N_{n=1}$ in observed space. Thus one could think of performing missing data reconstruction via dimensionality reduction. That is, if at some point $n$ in the sequence $\t_{\calM}$ are missing and $\t_{\calP}$ are present, we first reduce dimensionality by picking a representative point of $p(\x | \t_{\calP}) = \int{p(\x | \t) p(\t_{\calM} | \t_{\calP}) \, d\t_{\calM}}$ and then map that point onto observed space using the mapping \f. This will not work well except when $p(\x | \t_{\calP})$ is sharply unimodal, that is, $\t_{\calP}$ strongly constrains $\t_{\calM}$ to lie in a small region. But usually $\x | \t_{\calP}$ will be multimodal and therefore this is translating the problem of finding a multivalued relationship $\t_{\calP} \rightarrow \t_{\calM}$ to that of a multivalued dimensionality reduction $\t_{\calP} \rightarrow \x$! Besides, the dimensionality reduction approach will produce a value not just for $\t_{\calM}$ but also for $\t_{\calP}$, which may differ from the actually observed value of $\t_{\calP}$.

In fact,
\begin{equation*}
  p(\t_{\calM} | \t_{\calP}) = \int_{\calX}{p(\t_{\calM},\x | \t_{\calP}) \, d\x} = \int_{\calX}{p(\t_{\calM} | \x,\t_{\calP}) p(\x | \t_{\calP}) \, d\x} = \int_{\calX}{p(\t_{\calM} | \x) p(\x | \t_{\calP}) \, d\x}
\end{equation*}
where in the last equality we have used the fact that the observed variables are conditionally independent given the latent ones. Therefore, the procedure is equivalent to collapsing $\x | \t_{\calP}$ onto a delta function centred on $\hat{\x}$, throwing away all the probability mass not in $\hat{\x}$. For this same reason, in general it is not convenient to apply the continuity constraints to the latent variables rather than to the observed ones.

However, if we do want to reduce the dimensionality of a sequence with missing data, we can cast this as a reconstruction problem, where the missing variables are the latent variables \x\ and the present variables are the present observed variables $\t_{\calP}$. We can apply the ideas of this paper to the predictive distribution $p(\x | \t_{\calP})$.

\subsection{Underconstrained functions}
\label{s:discussion:underconstr}

When \y\ is underconstrained given \x, \y\ can take any value in a continuous manifold of dimension $Y \ge 1$, rather than taking values in a countable set (for $Y = 0$). This typically happens when there are too few present variables (at some point $n$ in the sequence). Geometrically, if the possible joint values of \x\ and \y\ span a manifold \calM\ of dimensionality $L$ in $\bbR^D$, then the set of possible values for \y\ given a fixed value $\x_0$ of \x\ is the intersection of \calM\ with the coordinate hyperplane $\x = \x_0$. The geometric approach has two disadvantages: it requires solving nonlinear systems of equations; and it disregards the noise, i.e., the stochastic variability of the data around and inside that manifold.

In the probabilistic framework the information about the data manifold \calM\ is embedded in the joint probability distribution $p(\t)$ of the observed variables, noise is taken care of and the only mathematical operations needed are conditioning (therefore marginalising) and finding all modes, which are computationally efficient for Gaussian mixtures. Using a Gaussian mixture as density model provides a partial but working solution to the underconstrained case, because the number of modes is finite if the Gaussian mixture is finite. Thus, the modes act as a finite sample of such manifold, and a quantisation error appears. This error can be reduced by using more mixture components, but at a cost that grows exponentially with the intrinsic dimensionality of the data.

In the extreme case where all variables are missing at point $n$, the modes are now the modes of the joint density function and can be computed once and stored for subsequent points where all variables are missing, to save computer time. If the density is a Gaussian mixture, another possibility with nil computational cost is simply to use all the component centroids, since in principle they should all lie in high-density areas of the data space. This will also produce a finer discretisation of the data manifold, since there should be fewer modes than centroids.

\subsection{Unbounded horizon problems}
\label{s:discussion:unbounded}

There are practical reconstruction problems where the data set to be reconstructed is infinite or long enough that the user periodically demands partial reconstruction; for example, in continuous speech with missing data, the user should receive reconstructed speech in real time, which requires that past speech be frozen once reconstructed, passed to the user and discarded for reconstruction of future speech. In operations research problems such as inventory control this is called an unbounded horizon problem.

The greedy algorithm requires no modification for unbounded horizon problems, but we do not recommend it for the reasons of section~\ref{s:constr:min-greedy}. The dynamic programming algorithm requires a finite sequence. A simple approach is to split the data stream into chunks (perhaps coinciding with user requests), reconstruct them separately and concatenate the reconstructed chunks. This has the risk of getting discontinuities at the splitting points and getting a suboptimal reconstruction of the whole stream, though for long enough chunks this may not be a problem.

Note that points where there is a unique pointwise reconstruction ($\nu_n = 1$) effectively split the layered graph into separate subgraphs (e.g.\ at node $n = 4$ in fig.~\ref{f:constr:min}). That is, whenever $\nu_n = 1$ the reconstructed trajectory for points earlier than $n$ can be frozen (to its optimal value) and the dynamic programming algorithm restarted from scratch, saving computer time and storage. Depending on the particular application and on the amount of missing data such zero-uncertainty points may be common; in speech, one likely example are silent frames, which are easily detected by thresholding the frame energy.

\subsection{Extensions}

Our approach has considered 1D constraints (i.e., sequential data). It would be interesting to consider \emph{multidimensional constraints}, e.g.\ reflecting spatial structure rather than (or as well as) temporal. An application where the experimental conditions are 2D is the recovery of wind fields from satellite scatterometer measurements \citep{Nabney_00a}. The strategy of section~\ref{s:constr:cont} of defining constraints based on the norm of finite difference approximations of derivatives can be readily generalised to multidimensional experimental conditions (see \citealp{CarreirGoodhil03a} for a related analysis in the context of elastic nets). An important problem with constraints of dimension $D$ higher than one is the curse of the dimensionality: the complexity of the multidimensional dynamic programming algorithm grows exponentially with $D$ \citep[pp.~141--143]{Durbin_98a}. Thus, global minimisation will not be feasible except for very small dimensions $D$. Further research is necessary to develop efficient heuristic approximations to multidimensional dynamic programming.


Our approach uses a continuity constraint. However, \emph{isolated discontinuities} may occur in sequences that are otherwise continuous (as a function of the experimental conditions). The discontinuities can be intrinsic to the nature of the problem or caused by undersampling. They pose challenging modelling difficulties, since they can confuse the dynamic programming search and cause a wrong global reconstruction. A possible approach is to use a robust local constraint, where the penalty saturates past a certain value of the reconstruction error.

\section{Related work}
\label{s:rel-work}

The key aspects of our approach are the use of a joint density model (learnt in an unsupervised way); the use of the modes of the conditional distribution as multiple pointwise candidate reconstructions; the mode search in Gaussian mixtures; the definition of a geometric trajectory measure derived from continuity constraints, and its minimisation by dynamic programming. Some of these ideas have been applied earlier in the literature in different contexts. However, we are not aware of any work that attempts to solve the missing data reconstruction problem in its full generality.

\subsection{Statistical approaches to missing data and imputation methods}
\label{s:rel-work:stat}

We have dealt with the problem ``given a data set with missing data, reconstruct it'' and we have assumed that a model for the data was available (perhaps obtained from a complete training set). The problem ``given a data set with missing data, estimate parameters (and standard errors, $p$-values, tests, etc.)\@ of a model of the data, or more generally, make inferences about the population from which the data come from'' has been the main concern of the statistical literature on missing data \citep{LittleRubin87,Little92,Schafer97a}. Such inferences must be done incorporating the missing data uncertainty; otherwise one will obtain too small standard errors. The pattern of missing data, given by the matrix \M\ of section~\ref{s:def:pat-md}, is considered a random variable. Calling $\T = \{\t_n\}^N_{n=1}$, etc., then the present data are $(\T_{\calP},\M)$ and the complete data $\T = (\T_{\calP},\T_{\calM},\M)$. If a joint distribution of $(\T,\M)$ with parameters \bTheta, \bPsi\ is assumed, $p(\T,\M|\bTheta,\bPsi) = p(\T|\bTheta) p(\M|\T,\bPsi)$, we are interested in inferences about \bTheta\ from $p(\bTheta,\bPsi|\T_{\calP},\M)$. The mechanism of missing data is usually assumed to be \emph{missing at random}: $p(\M|\T_{\calP},\T_{\calM},\bPsi) = p(\M|\T_{\calP},\bPsi)$ for all $\T_{\calM}$, i.e., the probability that a variable is missing does not depend on the value of that variable when it is missing.

The most popular methods are based on imputation, i.e., filling in the missing data and then estimating the parameters. The imputation can be single, usually the conditional mean given the present data; or, more effectively, multiple, where instead of imputing a single mean for each missing value, $M > 1$ values are drawn from the predictive distribution and then complete-data analyses repeated $M$ times, once with each imputation substituted, with the final parameter estimate being the average. Time-consuming Markov chain Monte Carlo methods are required.

In our method, we ignored any dependence between the probability that a variable be missing and the values that it or other variables may take. If information about such dependence was available, we could use it to further constrain the predictive distribution resulting in fewer candidate reconstructions. This would only be useful for varying missing data patterns, since we do not gain any information if it is always the same variables that are missing. Multiple imputation is then similar to the method of generating candidate reconstructions by sampling from the conditional distribution (section~\ref{s:deriv-funct-rel:sampling}), but there are major differences. In multiple imputation each imputation is done on the whole data set, not on each point separately; the latter would imply a number of imputations exponential (on the sample size). We avoided such an exponential complexity by minimising a constraint by dynamic programming. Also, the basic approach of multiple imputation and other statistical analysis methods for missing data consists of averaging over the missing data. This results in averaging branches of a multivalued mapping and contrasts with our method, which is based on mode finding and thus on branch identification.

\subsection{Multivalued mappings}
\label{s:rel-work:multiv-map}

Using a conditional distribution to define a mapping has been considered by other authors---in fact, the regression is defined in statistics by the conditional mean (under a given model) of the response given the predictor. For example, for function approximation, \citet{MoodyDarken89a} and \citet{Specht91a} used the mean of the conditional distribution derived from a kernel joint density estimate. For a classification problem with missing data and for function approximation, \citet{Ghahram94} and \citet{GhahramJordan94b} used a mixture model of the joint density and a single value from the conditional distribution: the mean, a random sample or the component centroid with highest probability%
\footnote{The ``component centroid with highest probability'' is often used as a fast approximation to the global mode of a mixture. However, it amounts to vector quantisation, since the selected value is always one of the centroids and all interaction between components is ignored. For Gaussian mixtures, the algorithms of \citet{Carreir00b} should indeed find the global mode as well as all the other modes.}.
\citet{Tresp_95a} used another method for regression with missing predictors based on a conditional mean. However, these approaches define a univalued mapping, in contrast to our proposal of using all the modes to define a multivalued mapping.

Other authors \citep{William86a,KinderLinden90a,Jensen_99a} have provided algorithms for inverting a trained neural net. If the latter (with fixed weights) implements a (forward) mapping $\y = \g(\x)$, then given a value \y\ any inverses of it must be local minima of the cost function $E(\x) = \norm{\y - \g(\x)}^2$. Gradient descent from different initial points can provide with some inverses, but one can never be sure of having found all them. Besides, not all local minima of $E$ are inverses, e.g.\ for $g(x) = x^3-x$ and $y = 2$, if starting with $x \le 0.5$ one gets $x \approx -0.58$ which is not an inverse. As mentioned in section~\ref{s:deriv-funct-rel:mean-harm}, the mode-finding algorithms of \citet{Carreir00b} return all the modes in most practical cases.

\subsection{Universal function approximators}
\label{s:rel-work:ua}

Multilayer perceptrons (MLPs) and other universal mapping approximators are excellent models to learn a univalued mapping, and if minimising the squared error they are generally equivalent to the conditional mean \citep{Bishop95a}. However, in our reconstruction problem, mapping approximators have two significant drawbacks. First, we need to model the mappings between many combinations of variables. Each combination requires a separate mapping approximator and the total number of combinations grows exponentially, also requiring sufficient training data for each combination. Second, we need a \emph{multivalued} mapping. A single mapping approximator results in a compromise mapping half way through the branches of the true mapping. We review some extensions of mapping approximators that have been proposed for mapping inversion. We consider the problem of approximating a multivalued mapping $\y \xrightarrow{\h} \x$ and will assume that it is the inverse of a univalued forward mapping $\x \xrightarrow{\g} \y$. None of the methods described here can deal with varying patterns of missing data.

\paragraph{Ensembles} Rather than solving a mapping approximation problem by using a single mapping approximator, one can use a finite collection of them, called an \emph{ensemble}. We need to represent every branch of the mapping with a different ensemble member: this achieves multiple pointwise reconstruction. A selection strategy can then be applied to attain a single reconstruction; constraint minimisation is an example. Several methods, proposed for articulatory inversion \citep{Rahim_93} and robot arm inverse kinematics \citep{DemersKreutz92,DemersKreutz96a,DemersKreutz98}, are based on the following tasks:
\begin{enumerate}
\item \emph{Branch determination}: this is the key part and requires to partition the space of the \x-variables into subsets over which the forward mapping is invertible. One can try to do this analytically only for the simplest problems; generally, one needs to cluster a training set (unsupervised learning).
\item \emph{Branch inversion}: the forward mapping restricted to a branch is by definition one-to-one, so a separate mapping approximator can now be fit by supervised learning to each branch to define a local inverse. The collection of all mappings, restricted to their respective subsets of \y-space, defines the ensemble and the global inverse.
\end{enumerate}
The advantage of these methods is that, while the learning stage (clustering and fitting the branches) is computationally costly, they are fast at run-time: inversion requires evaluating the local mapping approximators rather than mode finding. The disadvantage is that getting the right clustering is very difficult, particularly in high dimensions. This depends on heuristic, difficult-to-set parameters, such as neighbourhood sizes or cluster numbers (number of branches). Without a priori knowledge of the number of branches, it is difficult to detect when two clusters are really different. These parameters depend on the topologic and the geometric structure of the mapping manifold (e.g.\ its curvature) which is unknown in general. The clustering is sensitive to these parameters and a wrong clustering can seriously deteriorate the global inverse obtained. Thus, there is no guarantee that the local mappings are one-to-one inside every region and determining the regions is computationally costly in high dimensions.

The power of density estimation (admittedly difficult in high dimensions) is that it implicitly represents all the branches, i.e., implicitly determines the topology of the manifold. In our method, branch determination is achieved at reconstruction time by mode-finding in the corresponding conditional distribution.

A related method is the mixture-of-experts architecture \citep{Jacobs_91a,JordanJacobs94}, which is a set of function approximators (\emph{expert networks}) combined by a classifier (\emph{gating network}). The gating net, a multinomial logit model, softly splits the \x-space into regions where particular experts specialise but allowing data to be processed by multiple experts. The output of each expert is weighted by the gating network's estimate of the probability that that expert is the appropriate one to use, or a particular expert may be chosen according to the gating network's estimates, e.g.\ the one with the largest estimate. However, it is still restricted to learning \emph{univalued} mappings (in fact, \citealp{JordanJacobs94} applied it to the forward dynamics, not the inverse dynamics, of a robot arm).

\paragraph{Irreversible branch selection at training time} Direct application of a universal mapping approximator to a multivalued inverse mapping $\g^{-1}$ results in a univalued mapping \h\ equivalent to the conditional mean that may not be a solution for nonconvex problems, i.e., $\g(\h(\y)) \neq \y$. Several methods convert the multivalued mapping into a univalued one that is a valid inverse, i.e., that satisfies $\g(\h(\y)) = \y$. For example, \citet{JordanRumelh92} first train a neural network to model the forward mapping \g; then they prepend to it another network and retrain the resulting, cascaded network to learn the identity, $\y \rightarrow \y$, but keeping unchanged the weights of the forward model. This results in the prepended portion of the network learning one of the possible inverses. \citet{RohwerRest96} introduce a cost function with a description length interpretation whose minimum is approximated by the densest mode of a distribution. A neural network trained with this cost function can learn one branch of a multivariate mapping.

Therefore, these methods regularise the multivalued inverse mapping by adding some kind of constraints at training time so that the mapping becomes univalued: a single, particular branch is selected and the other inverses can never be recovered. However, trajectories that are contained in other branches will be incorrectly reconstructed, and the learned inverse mapping must contain discontinuous jumps between branches, similar to those of the global mode method (e.g.\ see fig.~\ref{f:exp-toy:results}E).

Note that the condition $\g(\h(\y)) = \y$ is the same as our forward-mapping constraint (section~\ref{s:constr:fwd-map}).

\paragraph{Recurrent nets} Feedforward nets, such as the MLP, are memoryless function approximators in that the predicted value depends only on the current input of a sequence. To represent information from the past, recurrent nets \citep{Hertz_91a,Robins94a,Tsoi98a} extend this architecture via feedback loops, e.g.\ to additional hidden units called context units or to a tapped delay line (time-delay neural networks). For observed data $\t^{(1)},\t^{(2)},\dots,\t^{(N)}$ they then estimate $\t^{(n)}|\t^{(1)},\dots,\t^{(n-1)}$, which makes them attractive for time series modelling.

Recurrent nets have a higher representational power than feedforward nets and they may conceivably be able to learn a constraint (given by the neighbouring sequence points) and an inverse mapping from data so that the right mapping branch is tracked at reconstruction time. However, they have the following drawbacks. (1) They are more difficult to train compared with feedforward nets (it requires large training sets, takes longer and there may be convergence problems) and do not generalise as reliably. (2) It may be difficult to find the right architecture for a given problem, particularly the number of context units or the time lag. This also applies to other time series models such as autoregressive models. (3) They fill in the data sequentially, like the greedy version of our algorithm, and so we would expect them to find suboptimal trajectories. 

\subsection{Conditional density modelling}
\label{s:rel-work:cond}

To learn a mapping $\y \xrightarrow{\h} \x$, one can use the conditional distribution $p(\x|\y)$ instead of a universal mapping approximator \citep{Bishop94a,William96a,Husmeier99a}. An example is the \emph{mixture density network} of \citet{Bishop94a} (see also \citealp[pp.~211--222]{Bishop95a}). This is a Gaussian mixture of spherical components whose parameters (means, variances, etc.)\ depend on the inputs \y\ through a mapping approximator, e.g.\ an MLP. \citet{Bishop94a} uses the centroid of one of the mixture components (e.g.\ that with highest mixture proportion) as an approximation to the global mode of the conditional distribution $\x|\y$. This results then in the same branch of the mapping being selected for a given value of \y\ (just as in the irreversible branch selection methods), with the rest of the information contained in the conditional distribution being ignored.

The conditional distribution obtained for a value of \y\ could be used to provide multiple pointwise reconstruction by finding its modes, as we propose in our method. And estimating only the conditional distribution is more efficient than estimating the joint density model, in view of the curse of the dimensionality. But, like function approximators, it treats the variables in an asymmetric way: \x\ missing and \y\ present. To reconstruct missing \y\ from present \x\ (for example) one would need the conditional distribution $p(\y|\x) \propto p(\x|\y) p(\y)$ which requires estimating the density of the \y\ variables or equivalently the joint density $p(\x,\y)$.

\subsection{Vector quantisation, codebooks and dynamic programming}
\label{s:rel-work:vq}

Consider again a known forward mapping $\g:\ \calX \rightarrow \calY$ to be inverted. In vector quantisation, one constructs a set of pairs $\{(\x_m,\y_m)\}^M_{m=1}\subset\calX\times\calY$ called \emph{codebook} where $\g(\x_m) = \y_m$ and the codebook thoroughly and finely spans the low-dimensional manifold of the mapping \g. At reconstruction time, given a point $\y\in\calY$, we search the codebook for candidate inverses that approximately map onto \y. There are different options for doing this:
\begin{enumerate}
\item \label{en:rel-work:vq:1} Return all codebook vectors $\x_m$ such that $d(\g(\x_m)-\y) < \epsilon$ where $d$ is a distance in the space \calY\ and $\epsilon$ is large enough to return at least one \x\ but not too large that many wrong $\x$s are returned.
\item \label{en:rel-work:vq:2} Return the $M'$ best inverses in order of distance $d(\g(\x_m)-\y)$, with $M' \ll M$.
\item \label{en:rel-work:vq:3} Return the whole codebook, i.e., $M' = M$.
\end{enumerate}
To invert a sequence $\{\y^{(n)}\}^N_{n=1}$, the pointwise candidate reconstructions provided by the codebook can be used with dynamic programming to minimise a cost function, such as continuity. Since the forward mapping is known, a constraint of the form of eq.~\eqref{e:constr:min:fwd} can be used too. Dynamic programming search of codebooks with options~\ref{en:rel-work:vq:2} or~\ref{en:rel-work:vq:3} has been used for articulatory inversion---the problem of finding the vocal tract configuration \x\ that produces an observed acoustic signal \y, a one-to-many mapping \citep{SchroetSondhi94a}. However, codebooks have the following disadvantages:
\begin{itemize}
\item Huge size: finely sampling in several dimensions implies very high codebook storage ($M > 100\,000$ entries for articulatory inversion) and search time.
\item Constructing the codebook is difficult. Among other reasons, simple clustering algorithms like $k$-means (where the means are the codebook vectors) cannot be used because the data manifold usually is not convex and so interpolated values may be illegal; and it is difficult to obtain a good sampling of the manifold because the forward mapping \g\ can stretch or compress the distance between neighbouring samples in the space \calX.
\item Even assuming a good codebook, the search returns fewer or, more likely, more inverse values than really exist (e.g.\ several per branch), which should result in the same problems as the spurious modes or the heuristic sampling of the conditional distribution (section~\ref{s:deriv-funct-rel:sampling}).
\item The reconstructed values are quantised, that is, only a finite number of different values is available to fill in \x, even though the range of \x\ is continuous.
\end{itemize}
Dynamic programming search of codebooks is a particular case of our method, the codebook being a zero-variance limit version of a mixture density model. The latter has the advantage of requiring many fewer parameters and (assuming a good density model) providing with the correct inverse values, without a neighbourhood parameter $\epsilon$ or $M'$, and, crucially, without quantisation artifacts (a continuous range is preserved for every variable).

\section{Conclusion}
\label{s:concl}

We have introduced the problem of reconstructing a sequence of vectors with missing data and given an algorithm for it. The algorithm exploits pointwise redundancy, in that the data are assumed to be intrinsically low-dimensional; and temporal redundancy, in that the sequence is assumed to vary smoothly. The algorithm works by first proposing at each vector in the sequence several candidate values for each missing variable; these values are given by the modes of a Gaussian-mixture conditional distribution (of the missing variables given the present ones). And secondly, the reconstructed sequence is obtained by minimising by dynamic programming a continuity constraint over all the candidates.

Our experiments have showed that the modes of the conditional distribution of the missing variables given the present ones are potentially plausible reconstructions of the missing values, and that the application of local continuity constraints---when they hold---can help to recover the actually plausible ones. We could call this approach \emph{modal regression} or \emph{modal reconstruction} (with constraints), in analogy with the standard definition of regression in terms of the mean of the conditional distribution of missing given present data.

\begin{figure}[t!]
  \begin{center}
    \begin{tabular}{@{}c@{\hspace{.1\textwidth}}c@{}}
      OFFLINE & AT RECONSTRUCTION TIME \\[1ex]
      \hline \\
      \begin{minipage}{.4\textwidth}
        \begin{center}
          \begin{tabular}{@{}c@{}}
            \caja{c}{c}{(Complete) dataset \\ $\{\t_n\}^{N'}_{n=1}$} \\[4ex]
            \includegraphics[height=.8cm]{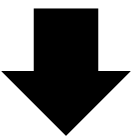} \\[1ex]
            \caja{c}{c}{Joint density model \\ estimation} \\[2ex]
            \fbox{\caja{c}{c}{\textbf{GTM} \\ \textbf{Gaussian mixture} \\ Kernel estimate \\ $\dots$ \\ (with regularisation \\ e.g.\ for smoothness)}} \\[10ex]
            \includegraphics[height=.8cm]{arrow-down} \\[1ex]
            \caja{c}{c}{Joint density model \\ $p(\t)$}
          \end{tabular}
        \end{center}
      \end{minipage} &
      \begin{minipage}{.4\textwidth}
        \begin{center}
          \begin{tabular}{@{}c@{}}
            \caja{c}{c}{Sequence $\{\t^{(n)}\}^N_{n=1}$ \\ with missing data} \\[3ex]
            \includegraphics[height=.8cm]{arrow-down} \\[1ex]
            \caja{c}{c}{Multiple pointwise reconstruction \\ from conditional distributions of $p(\t)$ \\ (for $n = 1,\dots,N$)} \\[4ex]
            \fbox{\caja{c}{c}{\textbf{Mode finding} \\ Bump finding \\ Random sample \\ $\dots$}} \\[7ex]
            \includegraphics[height=.8cm]{arrow-down} \\[1ex]
            \caja{c}{c}{Set of candidate \\ pointwise reconstructions} \\[4ex]
            \includegraphics[height=.8cm]{arrow-down} \\[1ex]
            \caja{c}{c}{Single global reconstruction \\ by constraint minimisation} \\[3ex]
            \fbox{\caja{c}{c}{\textbf{Dynamic programming} \\ Greedy algorithm \\ $\dots$ \\ with $\left\{ \mbox{\caja{c}{c}{\textbf{continuity constraint} \\ smoothness constraint \\ $\dots$}} \right.$}} \\[10ex]
            \includegraphics[height=.8cm]{arrow-down} \\[1ex]
            \caja{c}{c}{Reconstructed \\ sequence $\{\hat{\t}^{(n)}\}^N_{n=1}$}
          \end{tabular}
        \end{center}
      \end{minipage}
    \end{tabular}
  \end{center}
  \caption{Modular structure of the missing data reconstruction approach. The boxes represent modules that admit different implementations, such as the ones given; the ones recommended are in boldface. The density estimation stage (left) takes place offline.}
  \label{f:concl}
\end{figure}

Our method has the following characteristics:
\begin{itemize}
\item It is applicable to varying patterns of missing data: by using a joint probability model, the variables are treated symmetrically, unlike methods based on function approximators or conditional distribution approximators, which treat each variable either always as a predictor or always as a response. Predictive distributions for the missing data can be flexibly constructed as the corresponding conditional distribution.
\item It deals by design with multivalued mappings, representing all solution branches and choosing the right branch only at reconstruction time. This is unlike standard function approximators, which transform the multivalued mapping into a univalued one either by selecting always the same branch (irreversibly losing the others) or by averaging branches (which often results in a non-solution mapping).
\item By considering the pattern of missing data is constant, we can solve a mapping inversion problem. Here we need not model the joint density (always hard in high dimensions); we can simply model the conditional distribution of inputs given outputs. The inverse mapping can be constructed from measured input-output data, without knowledge of the functional form of the forward system---which can sometimes be difficult to obtain.
\item For general patterns of missing data, the method performs extremely robustly. For constant patterns of missing data (as in regression problems), it needs a reasonably smooth density model; otherwise the conditional distribution can contain spurious modes that result in suboptimal reconstructed trajectories with low constraint value. In mapping inversion problems, the effect of spurious modes may be eliminated by using a forward-mapping constraint.
\item It consists of several independent modules (fig.~\ref{f:concl}): joint density model, mode finding in conditional distributions and constraint minimisation by dynamic programming. Different algorithms, models or definitions may be used for each module.
\item It is insensitive to time warping, i.e., to reparametrisations of the trajectory, because the continuity constraint is the arc length---a geometric invariant.
\item It can also give confidence regions for each reconstructed value, derived from the Hessian at the corresponding mode that was selected.
\end{itemize}

The generality of our method means it can be applied to sequential data without the need to have an expert understanding of the problem, much like a neural net is applied to approximate an unknown function. Our method will be applicable to situations where multivalued relationships arise and interpoint constraints are available. This includes many mapping inversion problems, such as robot arm inverse kinematics and dynamics, estimation of 3D body pose and movement from a video sequence, articulatory inversion \citep{SchroetSondhi94a,CarreirRenals99} or decoding of neural population activity \citep{Zhang_98a}. An application where different variables may be missing at different times of the sequence is that of reconstructing occluded speech. A problem here is the definition of constraints, since acoustic features are in general not continuous. Perhaps perceptual grouping based on Gestalt principles, as used in computational auditory scene analysis \citep{BrownCooke94,CookeEllis01a}, could be helpful here. Another example is that of multimodal speech processing \citep{ChenRao98a}, where one wants to estimate the acoustics from the mouth shape (lip reading) and vice versa (facial animation) in the presence of occasional occlusion of either type of feature, with application to e.g.\ speech recognition. These are all hard problems because of the nonuniqueness of the (nonlinear) pointwise mappings and/or the variation of the pattern of missing data with time or space.

Generally speaking, our method is not applicable in the following cases. (1) Categorical variables: even though probability models can be constructed, the definition of local mode makes no sense, only the global mode does. (2) Independent data: if every data point $\t^{(n)}$ is independent of its neighbours $\t^{(n-1)}$, $\t^{(n+1)}$, etc.\ then no constraint across data points exists and consequently only multiple pointwise reconstruction is possible, not global reconstruction. Examples are i.i.d.\ data or shuffled data (where the original ordering of the data has been irreversible altered); such data sets are fine for training the joint pointwise density model, though. And, although it would work, the method should not be used as a replacement for universal mapping approximators (e.g.\ neural networks) in univalued mapping approximation problems (e.g.\ in forward mappings), since they are very efficient in this case.

\section{Acknowledgements}

We are grateful to Steve Renals for many helpful discussions.


\begin{thebibliography}{61}
\expandafter\ifx\csname natexlab\endcsname\relax\def\natexlab#1{#1}\fi
\expandafter\ifx\csname url\endcsname\relax
  \def\url#1{{\tt #1}}\fi

\bibitem[Asada and Slotine(1986)]{AsadaSlotin86a}
H.~Asada and J.-J.~E. Slotine.
\newblock {\em Robot Analysis and Control}.
\newblock John Wiley \& Sons, New York, London, Sydney, 1986.

\bibitem[Bellman(1957)]{Bellman57a}
R.~Bellman.
\newblock {\em Dynamic Programming}.
\newblock Princeton University Press, Princeton, 1957.

\bibitem[Bertsekas(1987)]{Bertsek87a}
D.~P. Bertsekas.
\newblock {\em Dynamic Programming. Deterministic and Stochastic Models}.
\newblock Prentice-Hall, Englewood Cliffs, N.J., 1987.

\bibitem[Bishop(1994)]{Bishop94a}
C.~M. Bishop.
\newblock Mixture density networks.
\newblock Technical Report \mbox{NCRG/94/004}, Neural Computing Research Group,
  Aston University, Feb. 1994.
\newblock Available online at
  \protect\url{http://www.ncrg.aston.ac.uk/Papers/postscript/NCRG_94_004.ps.Z}.

\bibitem[Bishop(1995)]{Bishop95a}
C.~M. Bishop.
\newblock {\em Neural Networks for Pattern Recognition}.
\newblock Oxford University Press, New York, Oxford, 1995.

\bibitem[Bishop et~al.(1998{\natexlab{a}})Bishop, Svens{\'e}n, and
  Williams]{Bishop_98b}
C.~M. Bishop, M.~Svens{\'e}n, and C.~K.~I. Williams.
\newblock Developments of the generative topographic mapping.
\newblock {\em Neurocomputing}, 21\penalty0 (1--3):\penalty0 203--224, Nov.
  1998{\natexlab{a}}.

\bibitem[Bishop et~al.(1998{\natexlab{b}})Bishop, Svens{\'e}n, and
  Williams]{Bishop_98a}
C.~M. Bishop, M.~Svens{\'e}n, and C.~K.~I. Williams.
\newblock {GTM}: The generative topographic mapping.
\newblock {\em Neural Computation}, 10\penalty0 (1):\penalty0 215--234, Jan.
  1998{\natexlab{b}}.

\bibitem[Brown and Cooke(1994)]{BrownCooke94}
G.~J. Brown and M.~Cooke.
\newblock Computational auditory scene analysis.
\newblock {\em Computer Speech and Language}, 8\penalty0 (4):\penalty0
  297--336, Oct. 1994.

\bibitem[Carreira-Perpi{\~n}{\'a}n(2000{\natexlab{a}})]{Carreir00b}
M.~{\'A}. Carreira-Perpi{\~n}{\'a}n.
\newblock Mode-finding for mixtures of {G}aussian distributions.
\newblock {\em \allcaps{IEEE} Trans. on Pattern Anal. and Machine Intel.},
  22\penalty0 (11):\penalty0 1318--1323, Nov. 2000{\natexlab{a}}.

\bibitem[Carreira-Perpi{\~n}{\'a}n(2000{\natexlab{b}})]{Carreir00a}
M.~{\'A}. Carreira-Perpi{\~n}{\'a}n.
\newblock Reconstruction of sequential data with probabilistic models and
  continuity constraints.
\newblock In S.~A. Solla, T.~K. Leen, and K.-R. M{\"u}ller, editors, {\em
  Advances in Neural Information Processing Systems}, volume~12, pages
  414--420. MIT Press, Cambridge, MA, 2000{\natexlab{b}}.

\bibitem[Carreira-Perpi{\~n}{\'a}n(2001)]{Carreir01a}
M.~{\'A}. Carreira-Perpi{\~n}{\'a}n.
\newblock {\em Continuous Latent Variable Models for Dimensionality Reduction
  and Sequential Data Reconstruction}.
\newblock PhD thesis, Dept. of Computer Science, University of Sheffield, UK,
  2001.
\newblock Available online at
  \protect\url{http://www.dcs.shef.ac.uk/~miguel/papers/phd-thesis.html}.

\bibitem[Carreira-Perpi{\~n}{\'a}n and Goodhill(2003)]{CarreirGoodhil03a}
M.~{\'A}. Carreira-Perpi{\~n}{\'a}n and G.~J. Goodhill.
\newblock Generalized elastic nets.
\newblock 2003.
\newblock Submitted.

\bibitem[Carreira-Perpi{\~n}{\'a}n and Renals(1999)]{CarreirRenals99}
M.~{\'A}. Carreira-Perpi{\~n}{\'a}n and S.~Renals.
\newblock A latent variable modelling approach to the acoustic-to-articulatory
  mapping problem.
\newblock In J.~J. Ohala, Y.~Hasegawa, M.~Ohala, D.~Granville, and A.~C.
  Bailey, editors, {\em Proc. of the 14th International Congress of Phonetic
  Sciences (ICPhS'99)}, pages 2013--2016, San Francisco, USA, Aug.~1--7 1999.

\bibitem[Carreira-Perpi{\~n}{\'a}n and
  Williams(2003{\natexlab{a}})]{CarreirWilliam03c}
M.~{\'A}. Carreira-Perpi{\~n}{\'a}n and C.~K.~I. Williams.
\newblock An isotropic {Gaussian} mixture can have more modes than components.
\newblock Technical Report \mbox{EDI--INF--RR--0185}, School of Informatics,
  University of Edinburgh, Dec. 2003{\natexlab{a}}.
\newblock Available online at
  \protect\url{http://www.informatics.ed.ac.uk/publications/report/0185.html}.

\bibitem[Carreira-Perpi{\~n}{\'a}n and
  Williams(2003{\natexlab{b}})]{CarreirWilliam03b}
M.~{\'A}. Carreira-Perpi{\~n}{\'a}n and C.~K.~I. Williams.
\newblock On the number of modes of a {Gaussian} mixture.
\newblock In L.~Griffin and M.~Lillholm, editors, {\em Scale Space Methods in
  Computer Vision}, volume 2695 of {\em Lecture Notes in Computer Science},
  pages 625--640, Berlin, 2003{\natexlab{b}}. Springer-Verlag.

\bibitem[Chen and Rao(1998)]{ChenRao98a}
T.~Chen and R.~R. Rao.
\newblock Audio-visual integration in multimodal communication.
\newblock {\em Proc. \allcaps{IEEE}}, 86\penalty0 (5):\penalty0 837--852, May
  1998.

\bibitem[Cooke and Ellis(2001)]{CookeEllis01a}
M.~Cooke and D.~P.~W. Ellis.
\newblock The auditory organization of speech and other sources in listeners
  and computational models.
\newblock {\em Speech Communication}, 35\penalty0 (3--4):\penalty0 141--177,
  Oct. 2001.

\bibitem[Cooke et~al.(2001)Cooke, Green, Josifovski, and Vizinho]{Cooke_01a}
M.~Cooke, P.~Green, L.~Josifovski, and A.~Vizinho.
\newblock Robust automatic speech recognition with missing and unreliable
  acoustic data.
\newblock {\em Speech Communication}, 34\penalty0 (3):\penalty0 267--285, June
  2001.

\bibitem[{DeGroot}(1986)]{Degroot86a}
M.~H. {DeGroot}.
\newblock {\em Probability and Statistics}.
\newblock Ad{\-d}i{\-s}on-Wes{\-l}ey, Reading, MA, 1986.

\bibitem[{DeMers} and Kreutz-Delgado(1992)]{DemersKreutz92}
D.~{DeMers} and K.~Kreutz-Delgado.
\newblock Learning global direct inverse kinematics.
\newblock In J.~Moody, S.~J. Hanson, and R.~P. Lippmann, editors, {\em Advances
  in Neural Information Processing Systems}, volume~4, pages 589--595. Morgan
  Kaufmann, San Mateo, 1992.

\bibitem[{DeMers} and Kreutz-Delgado(1996)]{DemersKreutz96a}
D.~{DeMers} and K.~Kreutz-Delgado.
\newblock Canonical parameterization of excess motor degrees of freedom with
  self-organizing maps.
\newblock {\em \allcaps{IEEE} Trans. Neural Networks}, 7\penalty0 (1):\penalty0
  43--55, Jan. 1996.

\bibitem[{DeMers} and Kreutz-Delgado(1998)]{DemersKreutz98}
D.~{DeMers} and K.~Kreutz-Delgado.
\newblock Learning global properties of nonredundant kinematic mappings.
\newblock {\em Int. J. of Robotics Research}, 17\penalty0 (5):\penalty0
  547--560, May 1998.

\bibitem[Durbin et~al.(1998)Durbin, Eddy, Krogh, and Mitchison]{Durbin_98a}
R.~Durbin, S.~R. Eddy, A.~Krogh, and G.~Mitchison, editors.
\newblock {\em Biological Sequence Analysis: Probabilistic Models of Proteins
  and Nucleic Acids}.
\newblock Cambridge University Press, 1998.

\bibitem[Durbin et~al.(1989)Durbin, Szeliski, and Yuille]{Durbin_89a}
R.~Durbin, R.~Szeliski, and A.~Yuille.
\newblock An analysis of the elastic net approach to the traveling salesman
  problem.
\newblock {\em Neural Computation}, 1\penalty0 (3):\penalty0 348--358, Fall
  1989.

\bibitem[Figueiredo and Jain(2002)]{FigueirJain02a}
M.~A.~T. Figueiredo and A.~K. Jain.
\newblock Unsupervised learning of finite mixture models.
\newblock {\em \allcaps{IEEE} Trans. on Pattern Anal. and Machine Intel.},
  24\penalty0 (3):\penalty0 381--396, Mar. 2002.

\bibitem[Ghahramani(1994)]{Ghahram94}
Z.~Ghahramani.
\newblock Solving inverse problems using an {EM} approach to density
  estimation.
\newblock In M.~C. Mozer, P.~Smolensky, D.~S. Touretzky, J.~L. Elman, and A.~S.
  Weigend, editors, {\em Proceedings of the 1993 Connectionist Models Summer
  School}, pages 316--323, 1994.

\bibitem[Ghahramani and Hinton(1996)]{GhahramHinton96a}
Z.~Ghahramani and G.~E. Hinton.
\newblock The {EM} algorithm for mixtures of factor analyzers.
\newblock Technical Report \mbox{CRG--TR--96--1}, University of Toronto, May~21
  1996.
\newblock Available online at
  \protect\url{ftp://ftp.cs.toronto.edu/pub/zoubin/tr-96-1.ps.gz}.

\bibitem[Ghahramani and Jordan(1994)]{GhahramJordan94b}
Z.~Ghahramani and M.~I. Jordan.
\newblock Supervised learning from incomplete data via an {EM} approach.
\newblock In J.~D. Cowan, G.~Tesauro, and J.~Alspector, editors, {\em Advances
  in Neural Information Processing Systems}, volume~6, pages 120--127. Morgan
  Kaufmann, San Mateo, 1994.

\bibitem[Harvey(1991)]{Harvey91a}
A.~C. Harvey.
\newblock {\em Forecasting, Structural Time Series Models and the {K}alman
  Filter}.
\newblock Cambridge University Press, 1991.

\bibitem[Hertz et~al.(1991)Hertz, Krogh, and Palmer]{Hertz_91a}
J.~A. Hertz, A.~S. Krogh, and R.~G. Palmer.
\newblock {\em Introduction to the Theory of Neural Computation}.
\newblock Ad{\-d}i{\-s}on-Wes{\-l}ey, Reading, MA, 1991.

\bibitem[Huber(1981)]{Huber81}
P.~J. Huber.
\newblock {\em Robust Statistics}.
\newblock John Wiley \& Sons, New York, London, Sydney, 1981.

\bibitem[Husmeier(1999)]{Husmeier99a}
D.~Husmeier.
\newblock {\em Neural Networks for Conditional Probability Estimation}.
\newblock Springer-Verlag, Berlin, 1999.

\bibitem[Jacobs et~al.(1991)Jacobs, Jordan, Nowlan, and Hinton]{Jacobs_91a}
R.~A. Jacobs, M.~I. Jordan, S.~J. Nowlan, and G.~E. Hinton.
\newblock Adaptive mixtures of local experts.
\newblock {\em Neural Computation}, 3\penalty0 (1):\penalty0 79--87, 1991.

\bibitem[Jensen et~al.(1999)Jensen, Reed, {Marks II}, {El-Sharkawi}, Jung,
  Miyamoto, Anderson, and Eggen]{Jensen_99a}
C.~A. Jensen, R.~D. Reed, R.~J. {Marks II}, M.~A. {El-Sharkawi}, J.-B. Jung,
  R.~T. Miyamoto, G.~M. Anderson, and C.~J. Eggen.
\newblock Inversion of feedforward neural networks: Algorithms and
  applications.
\newblock {\em Proc. \allcaps{IEEE}}, 87\penalty0 (9):\penalty0 1536--1549,
  Sept. 1999.

\bibitem[Jordan(1990)]{Jordan90}
M.~I. Jordan.
\newblock Motor learning and the degrees of freedom problem.
\newblock In M.~Jeannerod, editor, {\em Attention and Performance XIII}, pages
  796--836. Lawrence Erlbaum Associates, Hillsdale, New Jersey and London,
  1990.

\bibitem[Jordan and Jacobs(1994)]{JordanJacobs94}
M.~I. Jordan and R.~A. Jacobs.
\newblock Hierarchical mixtures of experts and the {EM} algorithm.
\newblock {\em Neural Computation}, 6\penalty0 (2):\penalty0 181--214, Mar.
  1994.

\bibitem[Jordan and Rumelhart(1992)]{JordanRumelh92}
M.~I. Jordan and D.~E. Rumelhart.
\newblock Forward models: Supervised learning with a distal teacher.
\newblock {\em Cognitive Science}, 16\penalty0 (3):\penalty0 307--354,
  July--Sept. 1992.

\bibitem[Kindermann and Linden(1990)]{KinderLinden90a}
J.~Kindermann and A.~Linden.
\newblock Inversion of neural networks by gradient descent.
\newblock {\em Parallel Computing}, 14\penalty0 (3):\penalty0 277--286, Aug.
  1990.

\bibitem[Lindeberg(1994)]{Lindeb94a}
T.~Lindeberg.
\newblock {\em Scale-Space Theory in Computer Vision}.
\newblock Kluwer Academic Publishers Group, Dordrecht, The Netherlands, 1994.

\bibitem[Little(1992)]{Little92}
R.~J.~A. Little.
\newblock Regression with missing {X's}: A review.
\newblock {\em J. Amer. Stat. Assoc.}, 87\penalty0 (420):\penalty0 1227--1237,
  Dec. 1992.

\bibitem[Little and Rubin(1987)]{LittleRubin87}
R.~J.~A. Little and D.~B. Rubin.
\newblock {\em Statistical Analysis with Missing Data}.
\newblock John Wiley \& Sons, New York, London, Sydney, 1987.

\bibitem[{McLachlan} and Krishnan(1997)]{MclachKrishn97a}
G.~J. {McLachlan} and T.~Krishnan.
\newblock {\em The {EM} Algorithm and Extensions}.
\newblock John Wiley \& Sons, New York, London, Sydney, 1997.

\bibitem[Moody and Darken(1989)]{MoodyDarken89a}
J.~Moody and C.~J. Darken.
\newblock Fast learning in networks of locally-tuned processing units.
\newblock {\em Neural Computation}, 1\penalty0 (2):\penalty0 281--294, Summer
  1989.

\bibitem[Nabney et~al.(2000)Nabney, Cornford, and Williams]{Nabney_00a}
I.~T. Nabney, D.~Cornford, and C.~K.~I. Williams.
\newblock {B}ayesian inference for wind field retrieval.
\newblock {\em Neurocomputing}, 30\penalty0 (1--4):\penalty0 3--11, Jan. 2000.

\bibitem[Nelson(1983)]{Nelson83a}
W.~L. Nelson.
\newblock Physical principles for economies of skilled movements.
\newblock {\em Biol. Cybern.}, 46\penalty0 (2):\penalty0 135--147, 1983.

\bibitem[Rabiner and Juang(1993)]{RabinerJuang93}
L.~Rabiner and B.-H. Juang.
\newblock {\em Fundamentals of Speech Recognition}.
\newblock Prentice-Hall, Englewood Cliffs, N.J., 1993.

\bibitem[Rahim et~al.(1993)Rahim, Goodyear, Kleijn, Schroeter, and
  Sondhi]{Rahim_93}
M.~G. Rahim, C.~C. Goodyear, W.~B. Kleijn, J.~Schroeter, and M.~M. Sondhi.
\newblock On the use of neural networks in articulatory speech synthesis.
\newblock {\em J. Acoustic Soc. Amer.}, 93\penalty0 (2):\penalty0 1109--1121,
  Feb. 1993.

\bibitem[Robinson(1994)]{Robins94a}
A.~J. Robinson.
\newblock An application of recurrent nets to phone probability estimation.
\newblock {\em \allcaps{IEEE} Trans. Neural Networks}, 5\penalty0 (2):\penalty0
  298--305, Mar. 1994.

\bibitem[Rohwer and van~der Rest(1996)]{RohwerRest96}
R.~Rohwer and J.~C. van~der Rest.
\newblock Minimum description length, regularization, and multimodal data.
\newblock {\em Neural Computation}, 8\penalty0 (3):\penalty0 595--609, Apr.
  1996.

\bibitem[Schafer(1997)]{Schafer97a}
J.~L. Schafer.
\newblock {\em Analysis of Incomplete Multivariate Data}.
\newblock Chapman \& Hall, London, New York, 1997.

\bibitem[Schroeter and Sondhi(1994)]{SchroetSondhi94a}
J.~Schroeter and M.~M. Sondhi.
\newblock Techniques for estimating vocal-tract shapes from the speech signal.
\newblock {\em \allcaps{IEEE} Trans. Speech and Audio Process.}, 2\penalty0
  (1):\penalty0 133--150, Jan. 1994.

\bibitem[Scott(1992)]{Scott92}
D.~W. Scott.
\newblock {\em Multivariate Density Estimation. Theory, Practice, and
  Visualization}.
\newblock John Wiley \& Sons, New York, London, Sydney, 1992.

\bibitem[Specht(1991)]{Specht91a}
D.~F. Specht.
\newblock A general regression neural network.
\newblock {\em \allcaps{IEEE} Trans. Neural Networks}, 2\penalty0 (6):\penalty0
  568--576, Nov. 1991.

\bibitem[Tikhonov and Arsenin(1977)]{TikhonArsenin77a}
A.~N. Tikhonov and V.~Y. Arsenin.
\newblock {\em Solutions of Ill-Posed Problems}.
\newblock John Wiley \& Sons, New York, London, Sydney, 1977.

\bibitem[Tipping and Bishop(1999)]{TippinBishop99a}
M.~E. Tipping and C.~M. Bishop.
\newblock Mixtures of probabilistic principal component analyzers.
\newblock {\em Neural Computation}, 11\penalty0 (2):\penalty0 443--482, Feb.
  1999.

\bibitem[Titterington et~al.(1985)Titterington, Smith, and Makov]{Titter_85a}
D.~M. Titterington, A.~F.~M. Smith, and U.~E. Makov.
\newblock {\em Statistical Analysis of Finite Mixture Distributions}.
\newblock John Wiley \& Sons, New York, London, Sydney, 1985.

\bibitem[Tresp et~al.(1995)Tresp, Neuneier, and Ahmad]{Tresp_95a}
V.~Tresp, R.~Neuneier, and S.~Ahmad.
\newblock Efficient methods for dealing with missing data in supervised
  learning.
\newblock In G.~Tesauro, D.~S. Touretzky, and T.~K. Leen, editors, {\em
  Advances in Neural Information Processing Systems}, volume~7, pages 689--696.
  MIT Press, Cambridge, MA, 1995.

\bibitem[Tsoi(1998)]{Tsoi98a}
A.~C. Tsoi.
\newblock Recurrent neural network architectures --- an overview.
\newblock In C.~L. Giles and M.~Gori, editors, {\em Adaptive Processing of
  Temporal Information}, volume 1387 of {\em Lecture Notes in Artificial
  Intelligence}, pages 1--26. Springer-Verlag, New York, 1998.

\bibitem[Williams(1996)]{William96a}
P.~M. Williams.
\newblock Using neural networks to model conditional multivariate densities.
\newblock {\em Neural Computation}, 8\penalty0 (4):\penalty0 843--854, May
  1996.

\bibitem[Williams(1986)]{William86a}
R.~J. Williams.
\newblock Inverting a connectionist network mapping by backpropagation of
  error.
\newblock In {\em Proc. 8th Annual Conf. Cognitive Science Society}, pages
  859--865. Lawrence Erlbaum, Hillsdale, NJ, 1986.

\bibitem[Zhang et~al.(1998)Zhang, Ginzburg, {McNaughton}, and
  Sejnowski]{Zhang_98a}
K.~Zhang, I.~Ginzburg, B.~L. {McNaughton}, and T.~J. Sejnowski.
\newblock Interpreting neuronal population activity by reconstruction: Unified
  framework with application to hippocampal place cells.
\newblock {\em J. Neurophysiol.}, 79\penalty0 (2):\penalty0 1017--1044, Feb.
  1998.

\end{thebibliography}
\ifx\undefined\allcaps\def\allcaps#1{#1}\fi

\end{document}